\documentclass{article}
\PassOptionsToPackage{numbers}{natbib}

\usepackage[preprint]{neurips_2026}
\usepackage[utf8]{inputenc} % allow utf-8 input
\usepackage[T1]{fontenc}    % use 8-bit T1 fonts
\usepackage{hyperref}       % hyperlinks
\usepackage{url}            % simple URL typesetting
\usepackage{booktabs}       % professional-quality tables
\usepackage{amsfonts}       % blackboard math symbols
\usepackage{microtype}      % microtypography
\usepackage[dvipsnames]{xcolor}
\usepackage{tcolorbox}
\usepackage{graphicx}       % figures

\newcommand{\kmedtablesymbol}{\raisebox{-0.12ex}{\includegraphics[height=0.8em]{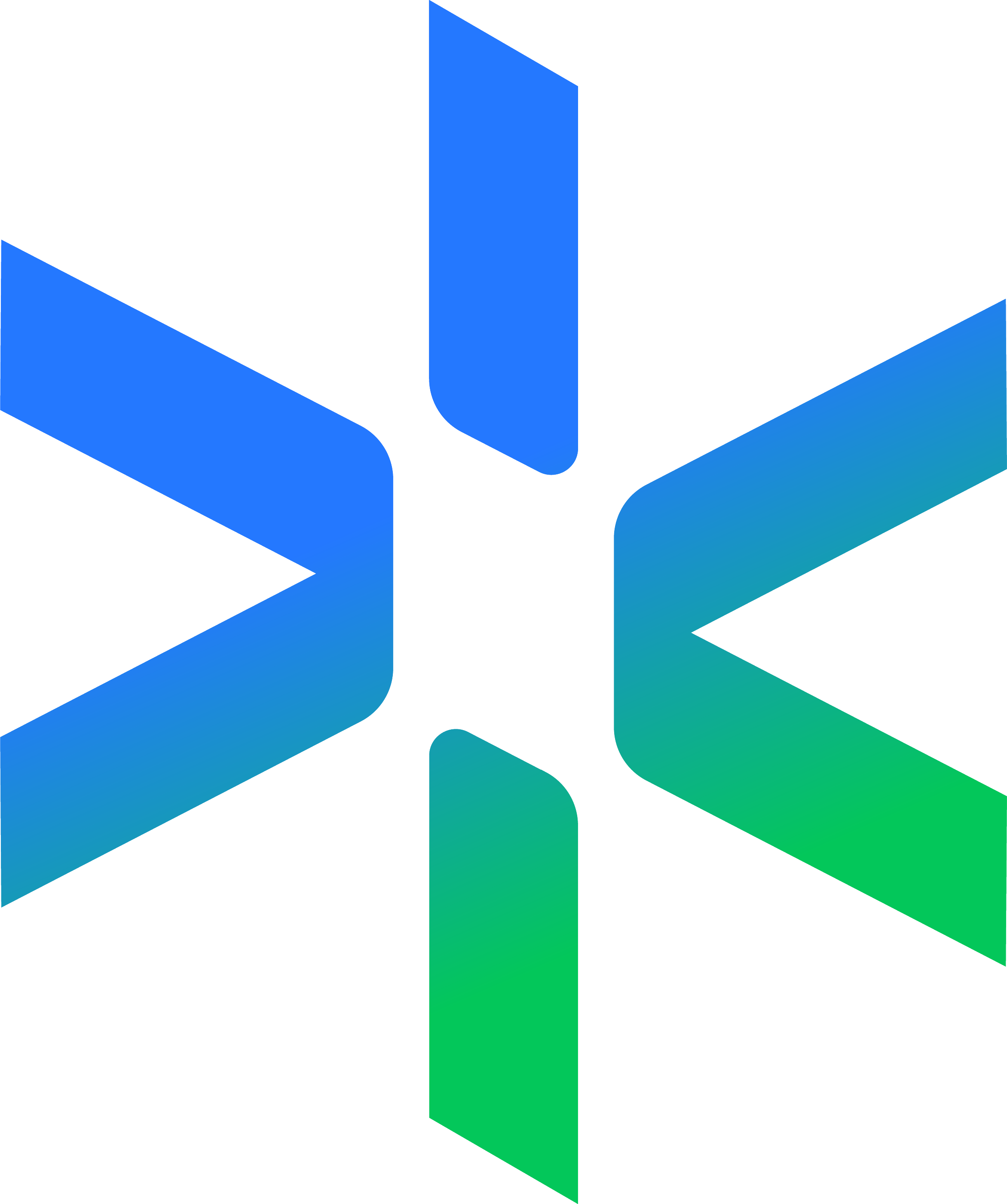}}\hspace{0.3em}}
\newcommand{\tabitem}{\hphantom{\kmedtablesymbol}}

\title{\raisebox{-0.28\height}{\includegraphics[height=22pt]{figs/kmed_symbol.png}}\hspace{7pt}KMed-31B Technical Report}

\author{KMed Team at NAVER}

\begin{document}

\maketitle

\begin{abstract}
We present KMed-31B, a medical large language model post-trained from an open-weight model for on-premises deployment with limited hardware. Post-training a model for diverse medical tasks remains difficult, because improving one kind of task can lower performance on another. We attribute this to differences between tasks in expected output and available feedback. Medical question answering and diagnosis prediction have reference answers, whereas clinical inquiry has no single correct response and is judged with rubrics. We therefore divide medical training tasks into four categories by these two properties, train one specialist per category with a reward that its feedback supports, and integrate the specialists into a single student with Multi-teacher On-policy Delta Distillation (MOPD$^2$).
For clinical inquiry, we build Synthetic Health Dialogue, a simulated corpus of consultations with lay users and clinicians whose patient cases span common clinical presentations in primary care. Relative to its base model, KMed-31B raises HealthBench Hard from 22.95 to 41.87 and HealthBench Professional from 37.92 to 50.20 in non-thinking mode, and from 23.80 to 51.48 and from 41.97 to 59.18 in thinking mode, while its scores on four multiple-choice medical benchmarks are maintained or slightly improved. KMed-31B has the highest unadjusted HealthBench Hard score among the compared models, including proprietary ones, and scores higher on HealthBench Professional than every open-weight model we compare, including models with more than ten times as many total parameters. Applied to another base model, Qwen3.8-27B, the same training recipe also raises both HealthBench scores. We describe the training recipe, data construction, and settings in detail so that institutions requiring on-premises deployment can develop medical models from open-weight foundations.
\end{abstract}

\section{Introduction}
\label{sec:introduction}

Frontier-level large language models have demonstrated strong performance on medical question answering and health conversations~\citep{arora2025healthbenchevaluatinglargelanguage,hicks2026healthbenchprofessional}.
In real clinical settings, however, services are often not allowed to send patient data outside the institution and must therefore run models on premises with limited hardware resources.
This requirement has motivated efforts to develop relatively small open-weight medical models~\citep{m2team2025baichuanm2scalingmedicalcapability,sellergren2026medgemmatechnicalreport}.
Despite their small size, these models must still answer knowledge-intensive medical questions, reason about clinical cases, and handle consultations with lay users and clinicians.

Recent open-weight models with around 30B total parameters match much larger models on individual medical benchmarks, but their strengths are uneven across multiple-choice medical questions and health conversations.
Gemma-4-31B-IT is on par with GPT-5.6-sol~\citep{openai2026gpt56systemcard} on MedXpertQA Text but falls far behind it on HealthBench Hard and HealthBench Professional, whereas Muse-Glimmer-30B~\citep{museglimmerteam2026museglimmer30b} approaches Qwen3.5-397B~\citep{qwenteam2026qwen35nativemultimodalagents} on both HealthBench benchmarks while its MedXpertQA Text score falls well below that of Gemma-4-31B-IT (Section~\ref{sec:general-comparison}).

\begin{figure*}[t]
\centering
\includegraphics[width=\textwidth]{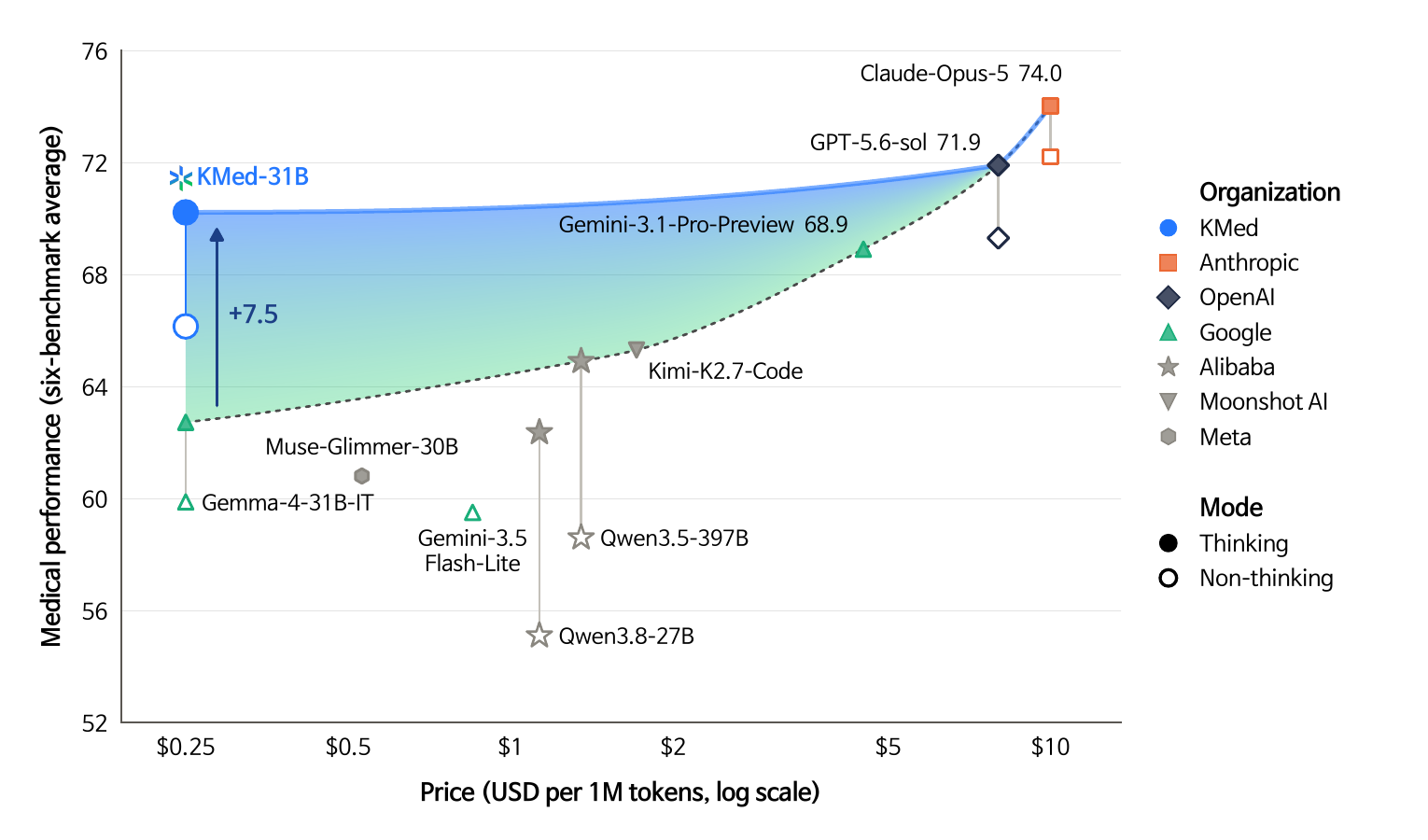}
\caption{Medical performance vs.\ price in thinking and non-thinking modes.
Medical performance is the mean of the scores on HealthBench Hard, HealthBench Professional, MedXpertQA Text, JAMA Clinical Challenge, NEJM AI QA, and MedQA-USMLE.
Blended prices use a 3:1 input-to-output token ratio.
We plot Gemma-4-31B-IT at \$0.25 per million tokens and assign the same price to KMed-31B.
For the other models, we use the list prices reported by Artificial Analysis~\citep{artificialanalysis_pricing} (first-party API prices when available, otherwise the median across providers).
Frontier curves are interpolated between Pareto-optimal points, and the arrow shows the thinking-mode increase from Gemma-4-31B-IT to KMed-31B.
}
\label{fig:headline-price-pareto}
\end{figure*}

\begin{figure*}[t]
\centering
\includegraphics[width=\textwidth]{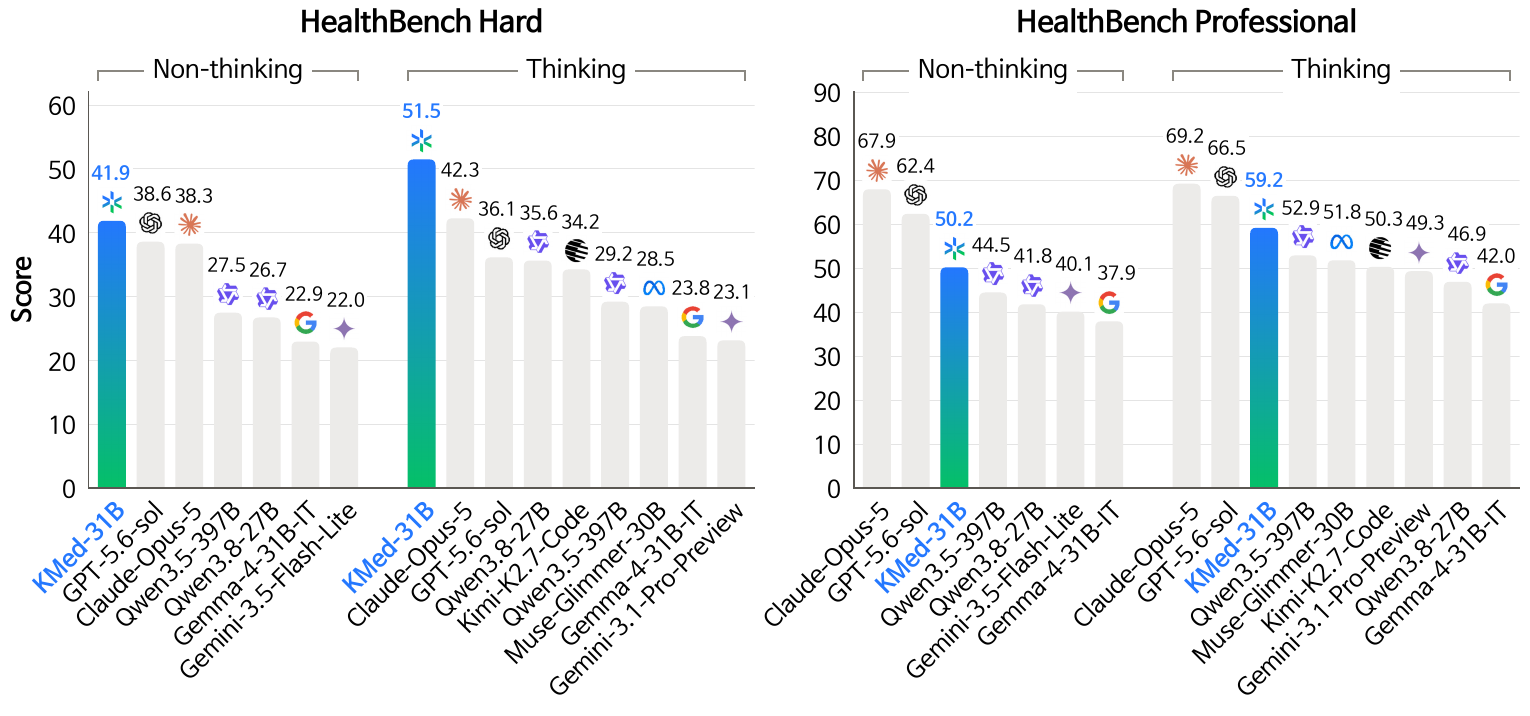}
\caption{Non-thinking and thinking results on HealthBench Hard (left) and HealthBench Professional (right), comparing KMed-31B with proprietary and open-weight models.
In both modes, KMed-31B has the highest HealthBench Hard score among the models shown and scores above all compared open-weight models on HealthBench Professional.
HealthBench scores are reported without length adjustment.}
\label{fig:headline-nonthinking}
\end{figure*}

Improving one of these capabilities through post-training can come at the expense of the other.
ConRub-Med reports that rubric-based reinforcement learning (RL) on Qwen3-32B raises HealthBench Hard from 9.46 to 33.17 while lowering MedXpertQA Text from 22.41 to 17.55~\citep{zhu2026conrubmedreinforcementlearningconsensus}.
In an Apertus-70B data ablation of Fully Open Meditron, adding synthetic clinical vignettes to the supervised fine-tuning (SFT) mixture improves open-ended clinical preference but lowers MedQA accuracy by 1.33 points~\citep{theimerlienhard2026fullyopenmeditronauditable}.
We observe a similar pattern when fine-tuning Gemma-4-31B-IT on consultation dialogues, which improves HealthBench performance but lowers MedXpertQA Text and JAMA Clinical Challenge accuracy even with medical multiple-choice examples mixed in (Section~\ref{sec:integration-analysis}).

The recurrence of this trade-off across base models and across both SFT and RL leads us to focus on how the tasks are posed and supervised rather than on a particular model or training method.
Medical question answering has reference answers against which a response can be checked, whereas clinical inquiry covers open-ended requests from lay users and clinicians that have no single correct response.
The quality of a clinical response depends on several aspects at once, such as its accuracy, its completeness, and its appropriateness to the user's context, and its evaluation therefore requires rubrics that specify criteria for each aspect.

Medical question answering and clinical inquiry thus differ in both expected output and available feedback.
In this sense, they constitute distinct domains, as mathematical reasoning and open-ended chat do in general-purpose post-training.
Such domains tend to interfere when trained together in a single run~\citep{cai2026advancinggeneralpurposereasoningmodels,wu2025imbalancedgradientsrlposttraining}, and each therefore calls for its own training procedure.
Prior medical post-training has addressed such interference by refining the training data mixture or by integrating separately trained specialists~\citep{ai2026cura1tspecializedmodel,m3team2026baichuanm3modelingclinicalinquiry}; we make this distinction explicit and use it to decide which specialists to train and how to train each one.

We develop KMed-31B from Gemma-4-31B-IT through a post-training recipe that groups medical tasks by expected output and available training feedback and assigns one specialist to each category (Section~\ref{sec:taxonomy}).
Open-ended question answering, multiple-choice question answering, and diagnosis prediction provide reference answers, so their specialists are trained through RL with answer-matching rewards.
Clinical inquiry has no single reference answer and relies on rubric-based feedback, so its specialist follows a dedicated recipe that proceeds from SFT to RL with fixed rubrics and then to Diversified ROPD (D-ROPD), our adaptation of rubric-based on-policy distillation (ROPD)~\citep{fang2026rubricbasedonpolicydistillation}.
In the final stage, we regenerate each query's rubric from fixed reference responses produced by several models together with the specialist's current rollouts (Section~\ref{sec:specialist-teachers}).

To obtain a single model for deployment, we integrate the specialists through Multi-teacher On-policy Delta Distillation (MOPD$^2$).
Multi-teacher on-policy distillation (MOPD) integrates separately trained domain specialists into a single student~\citep{mimo_v2_flash_2026,ma2026mopdmultiteacheronpolicydistillation}, and our MOPD$^2$ adopts the delta signal of on-policy delta distillation (OPD$^2$)~\citep{heo2026onpolicydeltadistillation}.
On the student's own rollouts, the specialist associated with each task category supplies a token-level signal based on its log-probability difference from a fixed reference model (the pretrained Gemma-4-31B checkpoint), so that the signal reflects what each specialist acquired relative to that checkpoint (Section~\ref{sec:mopd2}).

This recipe requires reference-answer datasets as well as clinical inquiry examples with demonstrations and rubric-based feedback.
We prepare the reference-answer tasks from public medical question-answering and clinical case datasets~\citep{jin2020diseasedoespatienthave,pal2022medmcqalargescalemultisubject,wu2025medcasereasoningevaluatinglearningdiagnostic}.
For clinical inquiry, we construct Synthetic Health Dialogue (SHD), a simulated corpus of consultations with lay users and clinicians whose patient cases are grounded in the clinical skills objectives of the Korean Medical Licensing Examination (Section~\ref{sec:synthesis}).
Its simulator separates the information available to the user from what the assistant can observe, creating conversations in which relevant context may initially be missing.
These conversations provide training queries for SFT and rubric-based RL, with demonstrations generated to encourage seeking missing context and checking the premises of a request.

In both thinking and non-thinking modes, KMed-31B achieves the highest unadjusted HealthBench Hard score among the compared models, including proprietary ones, and scores higher on HealthBench Professional than every open-weight model we compare, including models with more than ten times as many total parameters (Figure~\ref{fig:headline-nonthinking}).
Relative to its base model, KMed-31B raises HealthBench Hard from 22.95 to 41.87 and HealthBench Professional from 37.92 to 50.20 in non-thinking mode, without lowering accuracy on any of the four multiple-choice medical benchmarks (Section~\ref{sec:general-comparison}).
In thinking mode, it scores 51.48 on HealthBench Hard and 59.18 on HealthBench Professional, and it scores higher than the compared medical models from other developers with 27B to 32B parameters on every benchmark for which they have a score (Section~\ref{sec:medical-comparison}).
Applied to Qwen3.8-27B, the same training recipe also increases both HealthBench scores (Section~\ref{sec:architecture-analysis}).
With the per-token API price of its base model as a proxy, KMed-31B would extend the price-performance frontier of the comparison set for the six-benchmark average (Figure~\ref{fig:headline-price-pareto}).
We describe the training recipe, data construction, and settings in detail so that institutions requiring on-premises deployment can develop medical models from open-weight foundations.
\section{Task Taxonomy and Training Data}

\subsection{Task categories}
\label{sec:taxonomy}

We organize the training corpus around the task families of medical question answering, diagnosis prediction, and clinical inquiry (Table~\ref{tab:data-scale}).
The three families correspond to core capabilities of medical practice: (i) medical question answering trains the recall and application of medical knowledge, (ii) diagnosis prediction trains the identification of a diagnosis from a full case presentation, and (iii) clinical inquiry trains free-text responses to lay users and clinicians in conversation.
Together, they span the progression from medical knowledge to case-level diagnosis to communication with patients and clinicians.

Medical question answering is further divided by answer format into open-ended and multiple-choice items, which gives four task categories.
The categories differ in the form of the expected output and in the training feedback available: the two question-answering categories and diagnosis prediction have reference answers that support answer-matching rewards, whereas clinical inquiry responses have no single correct answer and are scored by a rubric-based judge.
For Gemma-4-31B-IT, we train a separate specialist teacher for each category in the non-thinking branch, while the thinking branch covers three of these task categories (Section~\ref{sec:training}).

\begin{figure}[t]
\centering
\includegraphics[width=\textwidth]{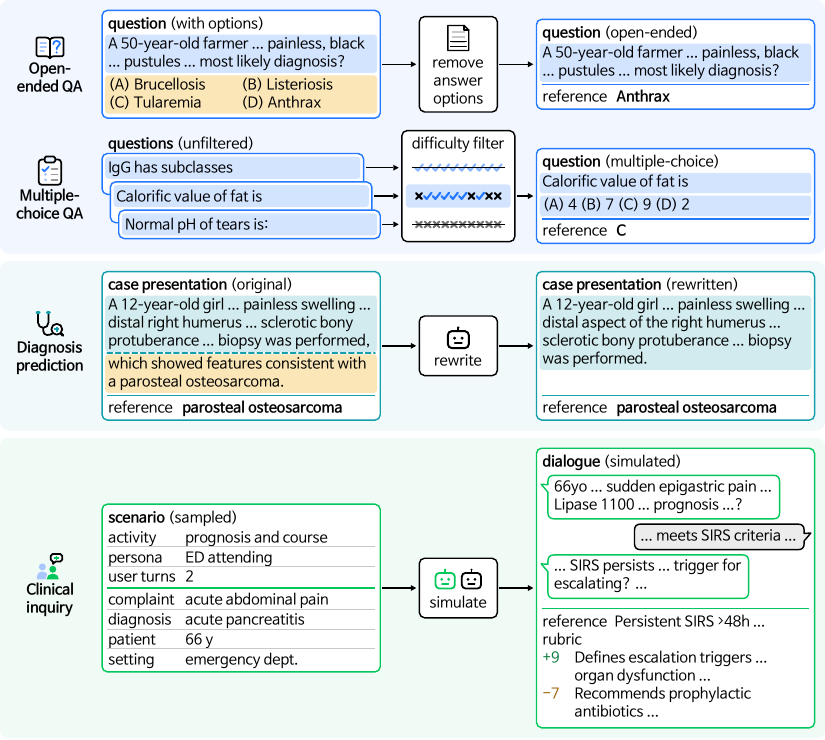}
\caption{Construction of the training data, with one example per task category.
From top to bottom, the examples come from MedQA-USMLE, MedMCQA, MedCaseReasoning, and SHD.
Each row shows the source (left), the construction step (middle), and the resulting training example (right).}
\label{fig:data-construction}
\end{figure}

We select or construct the data for each category as follows (Figure~\ref{fig:data-construction}).
For open-ended question answering, we use MedQA-OE, an open-ended version of the MedQA-USMLE training split~\citep{jin2020diseasedoespatienthave} in which the answer options are removed and the text of the correct option serves as the reference answer (Section~\ref{sec:medqa-oe-construction}).
Multiple-choice items are taken from a subset of the MedMCQA training split~\citep{pal2022medmcqalargescalemultisubject} that we select by difficulty.
The diagnosis prediction specialist is trained on the training split of MedCaseReasoning~\citep{wu2025medcasereasoningevaluatinglearningdiagnostic}, whose case presentations we rewrite to reduce explicit answer cues (Section~\ref{sec:mcr-rewriting}).
For clinical inquiry, we construct Synthetic Health Dialogue (SHD), a synthetic corpus of consultation exchanges with lay users and clinicians (Section~\ref{sec:synthesis}).

\begin{table}[t]
\centering
\small
\caption{Training data for the specialist teachers.
RLVR denotes reinforcement learning (RL) with verifiable rewards, in which the reward is a match against the reference answer.
Rubrics as Rewards (RaR) and D-ROPD are the two RL stages of the clinical inquiry specialist, with rubrics written in advance and with rubrics generated during training, respectively (Section~\ref{sec:specialist-teachers}).
All RL stages use GRPO (Section~\ref{sec:specialist-teachers}).
The \textit{\# Train examples} column provides the number of RL prompts, except for SFT, where it provides the number of query and response pairs used for supervised fine-tuning of the clinical inquiry specialist before RL.}
\label{tab:data-scale}
\setlength{\tabcolsep}{4pt}
\begin{tabular*}{\linewidth}{@{}l@{\extracolsep{\fill}}lllr@{}}
\toprule
\begin{tabular}[c]{@{}c@{}}Task\\category\end{tabular} & \begin{tabular}[c]{@{}c@{}}Training\\stage\end{tabular} & Source & Feedback & \begin{tabular}[c]{@{}c@{}}\# Train\\examples\end{tabular} \\
\midrule
Open-ended QA & RLVR & MedQA-USMLE (open-ended) & Answer match & 9.3k \\
Multiple-choice QA & RLVR & MedMCQA (filtered subset) & Answer match & 15.5k \\
Diagnosis prediction & RLVR & MedCaseReasoning (rewritten) & Answer match & 13k \\
Clinical inquiry & SFT & SHD (queries/responses) & Response supervision & 50k \\
Clinical inquiry & RL (RaR) & SHD (queries/rubrics) & Rubric judge & 10k \\
Clinical inquiry & RL (D-ROPD) & SHD (queries/reference responses) & Rubric judge & 2k \\
\bottomrule
\end{tabular*}
\end{table}

\subsection{Data construction}

\subsubsection{Medical question answering}
\label{sec:medqa-oe-construction}
\paragraph{Data sources and task formulation.}
Medical question answering uses examination questions with reference answers: MedQA-OE, converted from MedQA-USMLE~\citep{jin2020diseasedoespatienthave}, for the open-ended category and MedMCQA~\citep{pal2022medmcqalargescalemultisubject} for the multiple-choice category.
The two categories differ in the information available when answering.
MedMCQA presents answer options, whereas MedQA-OE requires the model to produce an answer without them.
We retain the two categories as separate specialist tasks rather than treating option selection and direct answer generation as interchangeable.

\paragraph{Open-ended conversion.}
Nemotron-CrossThink~\citep{akter2025nemotroncrossthink} removes answer options from multiple-choice training questions and reports higher accuracy on GPQA Diamond~\citep{rein2023gpqagraduatelevelgoogleproof} and SuperGPQA~\citep{du2025supergpqa} in its open-ended-format ablation.
Motivated by this result, we remove the answer options from MedQA-USMLE and use the text of the correct option as the reference answer, so that the specialist learns to generate the answer rather than select it from options.
The question text and source answer are retained; the conversion does not introduce newly generated answer labels.

\paragraph{Suitability screening and subset selection.}
Some MedQA-OE items depend on the original answer options or on visual material that is absent from the text.
We match the converted items to their source questions and apply an LLM-based suitability check to candidates flagged for visual or option dependence.
We screen these items and exclude those judged unsuitable for direct answering, including questions that require missing images or ECGs, questions that ask for an exception or the least likely option, and questions whose reference answer cannot be determined from the question alone.
An item is excluded only when the check judges it unsuitable, so candidates that the check marks as uncertain remain in the training set.
For MedMCQA, we use a difficulty-filtered subset that excludes questions that the initial checkpoint, Gemma-4-31B-IT, answered correctly in all or in none of 10 sampled responses.
This selection targets intermediate empirical difficulty; it does not guarantee non-zero reward variation under the evolving specialist policy.

\subsubsection{Diagnosis prediction}
\label{sec:mcr-rewriting}
\paragraph{Data source and task formulation.}
We use the MedCaseReasoning training split~\citep{wu2025medcasereasoningevaluatinglearningdiagnostic}, in which each case is derived from a published case report and consists of a case presentation, a reasoning section, and a final diagnosis label.
The specialist predicts the diagnosis from a case presentation rather than selecting from a supplied list of answer options.
The final diagnosis label serves as the reference for answer-matching rewards, but matching it does not by itself establish the completeness of a differential diagnosis.

\paragraph{Presentation rewriting.}
Some original MedCaseReasoning case presentations include the final diagnosis or the results of confirmatory tests.
To reduce these answer cues, we use Gemini-3.1-Pro-Preview to rewrite the case presentations and reasoning sections from the same source reports.
The prompt asks the model to retain presenting symptoms and initial findings while omitting diagnostic conclusions and confirmatory results from the presentation.
It also asks the model to support the reasoning section with quotations from the source report.
The source report is available during rewriting but is not appended to the specialist's case input.

\paragraph{Label preservation and quality checks.}
The rewritten training set preserves all 13,092 cases and their original diagnosis labels, which we use for reward computation.
We check the output schema and required fields, verify label preservation, and inspect diagnostic answer cues and the fidelity of quotations to the source reports, using automated checks and sampled model-based reviews.

\subsubsection{Clinical inquiry}
\label{sec:synthesis}
We generate Synthetic Health Dialogue (SHD) with an LLM-based simulator for two user populations: (i) lay users, who ask about their own health, the health of someone they care for, or health in general, and (ii) clinicians, who ask for help with patient care, clinical writing, and research.
The simulator aims to generate diverse and realistic dialogues across a wide range of clinical scenarios by modeling how these users converse with LLM-based assistants.

\paragraph{Clinical grounding.}
Patient cases for both populations come from a shared library based on the clinical skills examination objectives of the Korean Medical Licensing Examination~\citep{khplei2020cpx}.
These objectives define the minimum competencies expected of a primary care physician at licensure across 48 common clinical presentations and tasks.
The library covers all of these except breaking bad news and groups their possible causes or case types into 180 families based on the objectives' categories.
Each case combines one of these families with sampled patient attributes, which fix the diagnosis or clinical situation and the patient's demographic and clinical characteristics.

For lay users, we characterize conversation patterns using HealthChat-11K, which annotates 10,965 user conversations with general-purpose chatbots~\citep{paruchuri2025whatsupdoc}.
These patterns inform the distributions of 46 activities, user-turn counts, and follow-up actions.
For clinicians, we use the question-type frequencies reported by \citet{ely2000taxonomy} for primary care physicians to guide the distribution of patient-care requests.
For the specialty distribution of clinician personas, we use U.S. physician workforce statistics as a reference~\citep{aamc2024workforce,acgme2025databook}.
HealthChat-11K, Ely's questions, and the workforce statistics contribute only category-level counts, not individual conversations or questions.

\paragraph{Simulator design.}
We first use a planner to specify the conditions for each dialogue.
It samples the activity, the user persona, the number of user turns, and a patient case or topic.
It also assigns each follow-up turn an action, such as adding a detail, pushing back, or narrowing the request.
For lay users, it sets the first message's disclosure level (minimal, partial, or full), with probabilities informed by the context annotations on first messages in HealthChat-11K.
Given these conditions, Gemma-4-31B-IT generates user messages and assistant replies.

The model first prepares the user's initial information and opening message.
For lay users, a selected patient case is written as lived experience, with the diagnosis omitted unless the activity assumes it is known.
For clinicians, it is written as a chart in clinical shorthand, with added clinical details in a setting plausible for the persona's specialty.
The model also prepares facts the user may learn later, such as a new result or a change in symptoms.
Both the initial information and these later facts remain consistent with the sampled case.
Next, the model uses the planned activity, persona, and prepared case or selected topic to write the first message and a private record of the user's goal and relevant undisclosed information.
For lay users, the disclosure level is provided only when writing the first message.

The model then continues the dialogue by alternating between assistant replies and user follow-ups.
As in patient simulators for diagnostic dialogue~\citep{tu2024conversationaldiagnosticai,m2team2025baichuanm2scalingmedicalcapability}, the assistant sees only the dialogue so far, while the simulated user also receives the persona, prepared case or topic, private record, and assigned action.
At randomly selected assistant turns, the assistant is instructed to ask about missing information, except for lay users with full initial disclosure.
After such a turn, the user's next planned action may be replaced by answering the assistant.
For dialogues with a patient case, actions that add a detail, an update, or a new result introduce the next unused fact prepared earlier.
After each follow-up, information the simulated user reports as disclosed is removed from the private record.

\paragraph{Training queries, responses, and rubrics.}
Each simulated conversation ends with a user turn, and the full conversation serves as one training query for the next assistant response (Table~\ref{tab:data-scale}).
We drop a conversation whose full text or first user message repeats an earlier one.
For SFT, we use 50,000 clinician queries, pairing each with a reasoning trace and a final response from Kimi-K3 with thinking on.
A user's request may omit relevant information or rest on a mistaken premise.
These responses should therefore identify missing information that could materially change the answer, ask focused questions about it when needed, state assumptions without inventing patient-specific facts, and check the request's premises for consequential errors.
They are generated with a system prompt that states these requirements, which we call the context-and-premise prompt (Appendix~\ref{app:context-premise-prompt}).

For RL, we sample 5,000 lay queries and 5,000 of the 50,000 clinician queries.
Each RL query is also paired with a rubric that Kimi-K3 writes in advance from the conversation alone: a list of criteria specific to that conversation, each checkable from the conversation and a reply, with positive weights for what the best reply should do and negative weights for errors a reply might make.
RL prompts present a multi-turn query as a single message, and Section~\ref{sec:specialist-teachers} describes how rewards are computed in each RL stage.
\section{Training Recipe}
\label{sec:training}
\begin{figure}[t]
\centering
\includegraphics[width=\textwidth]{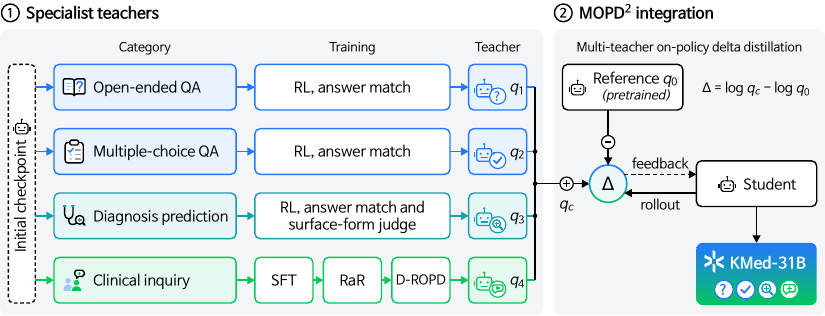}
\caption{Post-training pipeline for KMed-31B. 
Category-specific specialists are first trained with task-specific objectives and then integrated into a student through MOPD$^2$. 
For each prompt, the student is trained on its own rollouts using token-level feedback derived from the log-probability difference between the corresponding specialist $q_c$ and a pretrained reference model $q_0$. 
The pipeline is applied separately to the thinking and non-thinking branches (excluding open-ended QA in the thinking branch), yielding one checkpoint per branch.}
\label{fig:recipe-pipeline}
\end{figure}

Our post-training recipe has two stages (Figure~\ref{fig:recipe-pipeline}) and is defined by task categories, their rewards, and the integration procedure, without reference to a particular base model.
We apply it to Gemma-4-31B-IT and also to Qwen3.8-27B, to examine whether it carries over to another model family (Section~\ref{sec:architecture-analysis}).
First, we train one specialist teacher per task category, with a reward chosen according to the category's expected output and the feedback available for training (Section~\ref{sec:specialist-teachers}).
Each specialist keeps its own reward and training settings so that it is optimized for its target category and performs strongly within it.
Second, we apply OPD$^2$~\citep{heo2026onpolicydeltadistillation} to multi-teacher on-policy distillation (MOPD)~\citep{ma2026mopdmultiteacheronpolicydistillation}, called MOPD$^2$, to train a student to inherit the specialized teachers' performance in multiple domains (Section~\ref{sec:mopd2}).
For each model, the specialists and the student start from the same initial checkpoint, and only the student is used at inference.
The recipe has a non-thinking and a thinking branch, each with its own specialists and student.
For Gemma-4-31B-IT, the non-thinking branch covers all four task categories and the thinking branch covers all but open-ended question answering.
Qwen3.8-27B is trained only in the thinking branch, with all four categories.
Appendix~\ref{app:hyperparameters} lists the training settings of the non-thinking Gemma-4-31B-IT pipeline.

\subsection{Specialist teachers}
\label{sec:specialist-teachers}
Specialist teachers are trained using RL with verifiable rewards (RLVR) for the three categories with reference answers and rubric-based rewards for clinical inquiry (Section~\ref{sec:taxonomy}).
Training uses a leave-one-out variant of group relative policy optimization (GRPO)~\citep{shao2024deepseekmathpushinglimitsmathematical} on the RL prompts of each category (Table~\ref{tab:data-scale}).
For each prompt, the policy samples a group of $G$ responses, and the advantage of response $i$ with reward $R_i$ is $(R_i - \bar{R}_{-i})/\sigma_{-i}$, where $\bar{R}_{-i}$ and $\sigma_{-i}$ are the mean and standard deviation of the rewards other than $R_i$.
If the rewards other than $R_i$ are all equal, so that $\sigma_{-i} = 0$, the advantage is $R_i - \bar{R}_{-i}$.
The policy loss is a PPO-style clipped surrogate~\citep{schulman2017proximalpolicyoptimizationalgorithms} with a KL penalty toward the run's initial checkpoint, whose coefficient is set per category.

\paragraph{RLVR specialists.}
For the three answer-matching categories, we extract the final answer enclosed in answer tags and compare it with the reference answer after lowercasing and removing all characters other than letters and digits.
The reward is 1 on a match and 0 otherwise.
For diagnosis prediction only, an LLM judge (Gemma-4-31B-IT) reviews the responses that fail this match and assigns a reward of 1 when the two diagnoses differ in surface form alone.
A response truncated at the length limit receives 0.
In the answer-matching categories, groups whose responses all receive the same reward are excluded from the training batch, and additional prompts are sampled, following the dynamic sampling of DAPO~\citep{yu2025dapoopensourcellmreinforcement}.

\paragraph{Clinical inquiry specialist.}
Unlike the previous three tasks with clearly defined correct answers, clinical inquiry often involves open-ended questions with multiple clinically acceptable responses, requiring multi-criteria evaluation to assess response quality.
Thus, we introduce a three-stage training process to build the clinical inquiry specialist.
It starts with supervised fine-tuning (SFT), along with MedMCQA examples to retain multiple-choice performance.
Then, we conduct two learning stages based on rubrics: (i) RL with Rubrics as Rewards (RaR)~\citep{gunjal2025rubricsrewardsreinforcementlearning} and (ii) our Diversified ROPD (D-ROPD).
Here, a rubric defines a set of criteria that a high-quality response should satisfy for a given clinical inquiry, covering aspects such as clinical accuracy, completeness, and patient safety.

The clinical inquiry specialist training starts from SFT on Synthetic Health Dialogue (SHD) in Table~\ref{tab:data-scale}.
The SHD responses are generated with the context-and-premise prompt described in Section~\ref{sec:synthesis} (full text in Appendix~\ref{app:context-premise-prompt}).
The next stage is reinforcement learning based on RaR.
In RaR, we utilize rubrics written in advance for each query, and a judge reads the query, the rubric, and the final response and assigns one holistic rating, which is rescaled linearly to the unit interval as the reward, following the implicit variant of RaR~\citep{gunjal2025rubricsrewardsreinforcementlearning}.
The judge uses Gemma-4-31B-IT and does not receive a reference response.

\begin{figure}[t]
\centering
\includegraphics[width=\linewidth]{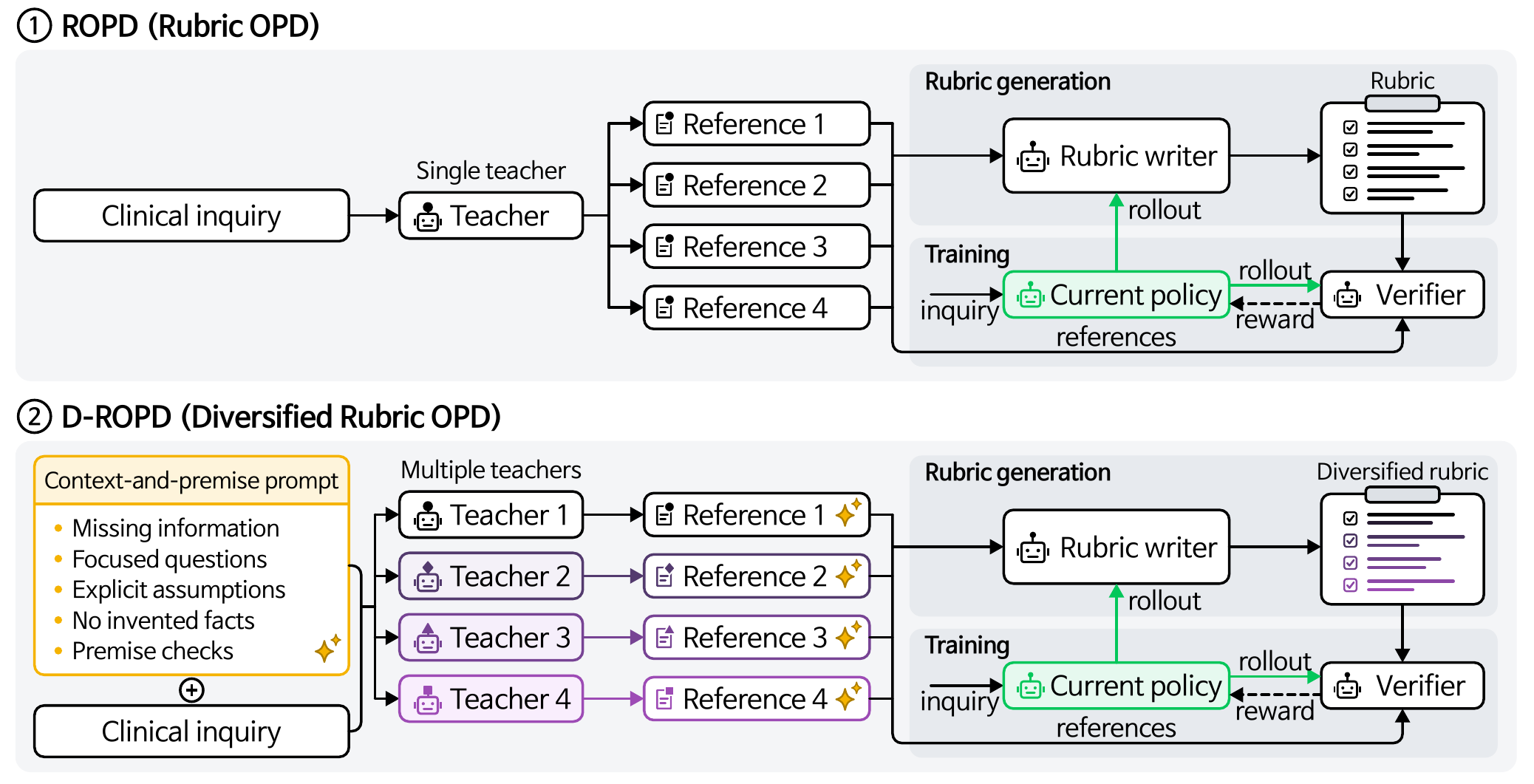}
\caption{Comparison of ROPD and D-ROPD in the final training stage of the clinical inquiry specialist.
ROPD samples several reference responses from one teacher model, whereas D-ROPD takes one reference response from each of four distinct teacher models, generated with the context-and-premise prompt.}
\label{fig:dropd}
\end{figure}

In the final stage, we introduce Diversified ROPD (D-ROPD), which extends rubric-based on-policy distillation (ROPD)~\citep{fang2026rubricbasedonpolicydistillation} from a single teacher to multiple teachers (Figure~\ref{fig:dropd}).
Unlike RaR, which relies on predefined rubrics, D-ROPD dynamically generates rubrics based on the current policy's responses and reference responses from diverse teachers.
By leveraging multiple teachers, D-ROPD aims to construct evaluation criteria that capture different valid approaches to the same clinical inquiry, rather than favoring a particular reference response.

Specifically, we use 2,000 clinical inquiry prompts from Table~\ref{tab:data-scale} and generate four reference responses for each query using distinct teacher models.
These reference responses are generated once using the same context-and-premise prompt.
For each training query, the rubric writer receives the reference responses with the current policy's rollouts and generates a set of binary evaluation criteria with positive integer weights.
Each criterion must be independently verifiable from a single response, without requiring similarity to the wording, structure, or final answer of any reference response.

To compute rewards, a verifier evaluates the current policy's rollouts against the generated rubric.
The verifier receives the query, rubric, and both the policy and reference responses in shuffled order without revealing their sources.
For each response, it determines whether each criterion is satisfied.
The reward is then computed as the weighted fraction of satisfied criteria:
\begin{equation}
r(y) = \frac{\sum_j w_j b_j(y)}{\sum_j w_j},
\end{equation}
where $w_j$ denotes the weight of criterion $j$, and $b_j(y) \in \{0,1\}$ indicates whether response $y$ satisfies the criterion.
Both the rubric writer and the verifier use Gemma-4-31B-IT.
Although reference responses are used for rubric generation and verification, they are excluded from the GRPO group and reward statistics.

\subsection{Multi-teacher OPD$^2$ (MOPD$^2$)}
\label{sec:mopd2}
We integrate the specialists trained in Section~\ref{sec:specialist-teachers} into one student by using Multi-teacher OPD$^2$ (MOPD$^2$), which applies the delta signal of on-policy delta distillation (OPD$^2$)~\citep{heo2026onpolicydeltadistillation} to multi-teacher on-policy distillation (MOPD)~\citep{ma2026mopdmultiteacheronpolicydistillation}.
Although each specialist demonstrates improved performance on its own domain, it is hard to integrate multi-domain RL settings into one model.
A feasible option is to use multi-domain RL with carefully tuned domain balance, yet we empirically observe that it is not simple to achieve multi-domain performance with a multi-domain RL framework.
Thus, we employ MOPD based on the improved objective OPD$^2$, which we call Multi-teacher OPD$^2$ (MOPD$^2$).
Instead of the difference between teacher and student, it computes the log-probability difference between the specialist teacher and its base model, which inherits the knowledge obtained for specialist training.
Our OPD$^2$ implementation follows \citet{heo2026onpolicydeltadistillation} and uses the model before instruction tuning (e.g., Gemma-4-31B) as the teacher base.

\paragraph{Training setup.}
During MOPD$^2$, the specialist teacher models and the corresponding teacher base model $q_0$ stay frozen while the student is updated on equally proportioned multi-domain training batches.
The prompts come from the RL training data of each category.
We group prompts into category-specific batches, and assign each batch to the corresponding specialist teacher.
Each batch uses 512 prompts with one rollout per prompt, and training is conducted for a total of 100 iterations.

\paragraph{Token advantage.}
Let $y$ be the student's rollout for a prompt $x$ of category $c$.
$q_c$ is the specialist teacher for category $c$ and $s_t = (x, y_{<t})$ is the context of token $y_t$.
$\mathbb{E}_t$ denotes the expectation under the student's next-token distribution $\pi_\theta(v \mid s_t)$.
With the delta $\Delta_t(v) = \log q_c(v \mid s_t) - \log q_0(v \mid s_t)$ and the distillation difference $D_t(v) = \log q_c(v \mid s_t) - \log \pi_\theta(v \mid s_t)$, OPD$^2$ centers both under the student and forms the advantage~\citep{heo2026onpolicydeltadistillation}
\begin{equation}
A^{\Delta}_t = \Delta_t(y_t) - \mathbb{E}_t\left[\Delta_t(v)\right], \qquad A^{\mathrm{OPD}}_t = D_t(y_t) - \mathbb{E}_t\left[D_t(v)\right],
\label{eq:mopd2-signals}
\end{equation}
\begin{equation}
\hat{A}_t = A^{\Delta}_t \; \mathbb{I}\left\{A^{\Delta}_t A^{\mathrm{OPD}}_t > 0\right\}.
\label{eq:mopd2-advantage}
\end{equation}
The advantage value is the centered delta, except when the delta signal is opposite to the original OPD signal $A^{\mathrm{OPD}}_t$.

\paragraph{Objective.}
For advantage centering, we approximate the expectation by a sum over the top-64 vocabulary entries, rather than the 1,024 used by OPD$^2$~\citep{heo2026onpolicydeltadistillation}.
When the sampled token is not in the top-64 vocabulary, it replaces the least probable entry.
The student then minimizes
\begin{equation}
\mathcal{L}(\theta) = -\frac{1}{|\mathcal{B}|} \sum_{y \in \mathcal{B}} \frac{1}{|y|} \sum_{t=1}^{|y|} \hat{A}_t \log \pi_\theta(y_t \mid s_t)
\label{eq:mopd2-loss}
\end{equation}
over a batch $\mathcal{B}$ of rollouts, holding $\hat{A}_t$ fixed.
We also use the KL divergence from the reference model as a regularizer for MOPD$^2$ training, similar to the RL recipe.
For models that do not provide a pretrained base model, such as Qwen3.8-27B, we use a base model from an earlier version with a similar size (e.g., Qwen3.5-35B-A3B-Base) as the reference model $q_0$ in the OPD$^2$ objective.
\section{Evaluation}
\label{sec:evaluation}

\subsection{Evaluation setup}
\label{sec:eval-setup}

\paragraph{Benchmarks.}
We evaluate KMed-31B on six medical benchmarks covering health conversations, clinical case reasoning, and licensing and specialty examination questions.
Two of them, HealthBench Hard and HealthBench Professional, evaluate free-text responses using physician-written rubrics.
HealthBench Hard is a subset of 1,000 examples, selected as difficult even for frontier models, from HealthBench~\citep{arora2025healthbenchevaluatinglargelanguage}, which contains 5,000 conversations between a model and a lay user or a health professional.
HealthBench Professional~\citep{hicks2026healthbenchprofessional} consists of 525 conversations that physicians wrote while testing a chat assistant for clinicians, about one third of them red-teaming attempts.
In both, the rubrics assess the accuracy, completeness, and contextual appropriateness of the next assistant response.

The remaining four are multiple-choice benchmarks, evaluated with text-only inputs.
From MedXpertQA~\citep{zuo2025medxpertqabenchmarkingexpertlevelmedical} we use the text subset, MedXpertQA Text, which contains difficult questions drawn from licensing and specialty examinations.
JAMA Clinical Challenge~\citep{chen2025benchmarkinglargelanguagemodels} asks for the diagnosis or the next management step in published clinical cases.
NEJM AI QA~\citep{katz2024gptresidentphysicians} contains English translations of multiple-choice questions from the 2022 Israeli specialist board examinations.
MedQA-USMLE~\citep{jin2020diseasedoespatienthave} contains clinical questions in the style of the U.S. medical licensing examination.

\paragraph{Compared models.}
We compare KMed-31B with general-purpose open-weight models spanning a wide range of parameter counts, proprietary models accessed through APIs, and models specialized for the medical domain by other developers.
The non-thinking and thinking versions of KMed-31B are trained separately (Section~\ref{sec:training}).
In the comparison with general-purpose models, we report the two versions separately and compare each with models evaluated in the corresponding mode, since generating reasoning traces increases output length and thereby inference latency and cost.
The medical models include both open-weight and closed-weight models of varying sizes, and we indicate whether each uses thinking mode.
We generate responses from KMed-31B using temperature $1.0$, top-$p$ $0.95$, and top-$k$ $64$, while the other models evaluated under our protocol use the default generation settings of their respective providers or serving configurations.

\paragraph{HealthBench scoring.}
For HealthBench Hard and HealthBench Professional, each model receives the conversation provided in each example and generates the next assistant response.
Each final assistant response is judged by GPT-5-mini, which marks every physician-written criterion as met or not met.
Following the HealthBench scoring rule~\citep{arora2025healthbenchevaluatinglargelanguage}, an example's score is the sum of the weights of the met criteria, including negative weights, divided by the sum of all positive weights.
Within each run, we average the example scores, clip the mean to $[0,1]$, and multiply it by 100.
The original HealthBench and HealthBench Professional reports use GPT-4.1 and GPT-5.4 as graders, respectively, so our scores are not directly comparable with the scores published there.

Longer responses tend to meet more criteria, and Appendix~\ref{app:length-adjusted} reports the length-adjusted score defined for HealthBench Professional~\citep{hicks2026healthbenchprofessional}.
All HealthBench scores in this section are reported without length adjustment.
We write $\mathrm{avg}@k$ for the mean of $k$ run-level scores, where each run generates and grades new responses to all examples, and report HealthBench scores as avg@3.

\paragraph{Multiple-choice scoring.}
We evaluate the multiple-choice benchmarks using EvalScope~\citep{evalscope_2024}.
We report accuracy (\%) as avg@10, based on answers extracted from the final responses.
NEJM AI QA includes items with multiple reference answers; we count an extracted answer as correct only if it exactly matches the full reference answer set.

\begin{table}[t]
\centering
\small
\caption{Comparison with general-purpose models in non-thinking and thinking modes.}
\label{tab:general-comparison}
\setlength{\tabcolsep}{4pt}
\renewcommand{\arraystretch}{1.06}
\begin{tabular*}{\linewidth}{@{}l@{\extracolsep{\fill}}cccccc@{}}
\toprule
Model & \begin{tabular}[c]{@{}c@{}}HealthBench\\Hard\end{tabular} & \begin{tabular}[c]{@{}c@{}}HealthBench\\Pro.$^{*}$\end{tabular} & \begin{tabular}[c]{@{}c@{}}MedXpertQA\\Text\end{tabular} & \begin{tabular}[c]{@{}c@{}}JAMA\\C. C.$^{\dagger}$\end{tabular} & \begin{tabular}[c]{@{}c@{}}NEJM\\AI QA\end{tabular} & \begin{tabular}[c]{@{}c@{}}MedQA-\\USMLE\end{tabular} \\
\midrule
\multicolumn{7}{@{}l@{}}{\textit{Non-thinking, proprietary models}} \\
\tabitem Claude-Opus-5 & 38.29 & \textbf{67.94} & \textbf{60.45} & \textbf{87.20} & \textbf{83.63} & \textbf{95.85} \\
\tabitem GPT-5.6-sol & \textbf{38.59} & 62.36 & 52.55 & 86.41 & 82.39 & 93.75 \\
\tabitem Gemini-3.5-Flash-Lite & 22.02 & 40.13 & 43.80 & 78.01 & 80.62 & 92.51 \\
\addlinespace[3pt]
\multicolumn{7}{@{}l@{}}{\textit{Non-thinking, open-weight models and KMed-31B}} \\
\tabitem GLM-5.2 (753B) & 21.29 & 41.39 & 30.03 & 76.06 & 70.74 & 84.66 \\
\tabitem Qwen3.5-397B & 27.47 & 44.47 & 35.88 & \textbf{77.94} & 75.91 & 90.10 \\
\tabitem Gemma-4-31B-IT & 22.95 & 37.92 & 53.86 & 75.89 & 75.54 & 93.08 \\
\tabitem Qwen3.8-27B & 26.72 & 41.79 & 33.99 & 74.08 & 67.92 & 85.91 \\
\kmedtablesymbol \textbf{KMed-31B} & \textbf{41.87} & \textbf{50.20} & \textbf{56.32} & 76.06 & \textbf{78.03} & \textbf{94.38} \\
\midrule
\multicolumn{7}{@{}l@{}}{\textit{Thinking, proprietary models}} \\
\tabitem Claude-Opus-5 & \textbf{42.26} & \textbf{69.20} & 62.29 & \textbf{88.88} & \textbf{85.13} & 96.33 \\
\tabitem GPT-5.6-sol & 36.14 & 66.45 & 59.51 & 87.71 & 85.08 & 96.41 \\
\tabitem Gemini-3.1-Pro-Preview & 23.14 & 49.32 & \textbf{71.27} & 87.80 & 85.03 & \textbf{96.54} \\
\addlinespace[3pt]
\multicolumn{7}{@{}l@{}}{\textit{Thinking, open-weight models and KMed-31B}} \\
\tabitem Kimi-K2.7-Code (1T) & 34.25 & 50.30 & 48.11 & 82.96 & 81.43 & 94.81 \\
\tabitem GLM-5.2 (753B) & 23.23 & 46.34 & 51.42 & \textbf{82.99} & \textbf{81.59} & \textbf{95.21} \\
\tabitem Qwen3.5-397B & 29.19 & 52.90 & 51.11 & 79.70 & 81.38 & 95.11 \\
\tabitem Gemma-4-31B-IT & 23.80 & 41.97 & 58.62 & 78.02 & 79.49 & 94.41 \\
\tabitem Muse-Glimmer-30B & 28.46 & 51.80 & 38.57 & 76.98 & 76.84 & 92.33 \\
\tabitem Qwen3.8-27B & 35.64 & 46.90 & 43.95 & 77.94 & 75.45 & 94.27 \\
\kmedtablesymbol \textbf{KMed-31B} & \textbf{51.48} & \textbf{59.18} & \textbf{58.64} & 77.86 & 79.75 & 94.38 \\
\bottomrule
\end{tabular*}
\vspace{3pt}
\begin{minipage}{\linewidth}
\footnotesize
$^{*}$HealthBench Professional.
$^{\dagger}$JAMA Clinical Challenge.
HealthBench columns are avg@3 and multiple-choice columns are avg@10.
Bold marks the best score in each column within each group.
\end{minipage}
\end{table}

\subsection{Comparison with general-purpose models}
\label{sec:general-comparison}

Table~\ref{tab:general-comparison} compares KMed-31B with proprietary and open-weight general-purpose models, and Figure~\ref{fig:headline-nonthinking} shows the scores on the two HealthBench benchmarks.
Compared with the proprietary models, KMed-31B scores highest in both modes on HealthBench Hard, the HealthBench subset selected as difficult for frontier models, ahead of GPT-5.6-sol (38.59) in non-thinking mode and Claude-Opus-5 (42.26) in thinking mode.
On HealthBench Professional, which reflects real-world clinician requests, it scores above the Gemini model evaluated in each mode and below GPT-5.6-sol and Claude-Opus-5.
Its difference from Claude-Opus-5 on this benchmark is 17.74 points in non-thinking mode and 10.02 points in thinking mode, compared with 30.02 and 27.23 points for Gemma-4-31B-IT.

Compared with the open-weight models, KMed-31B scores highest on both HealthBench Hard and HealthBench Professional in both modes, although several of these models have more than ten times as many total parameters.
On HealthBench Hard, it scores 41.87 in non-thinking mode and 51.48 in thinking mode, and the next highest scores are 27.47 for Qwen3.5-397B and 35.64 for Qwen3.8-27B.
On HealthBench Professional, it scores 50.20 and 59.18, and the next highest scores are 44.47 and 52.90, both for Qwen3.5-397B.

Relative to Gemma-4-31B-IT, KMed-31B gains between 12.28 and 27.68 points on the two health conversation benchmarks across the two modes, while no score on the four multiple-choice benchmarks of examination questions and clinical cases decreases by more than 0.16 points (Section~\ref{sec:thinking-gap}).
Among the other compared models, a high score on examination questions is not always accompanied by a high score on health conversations.
For instance, Gemini-3.1-Pro-Preview in thinking mode scores highest on the difficult examination questions of MedXpertQA Text, at 71.27, but scores 23.14 on HealthBench Hard.

KMed-31B remains below the proprietary models in both modes on JAMA Clinical Challenge, which asks for the diagnosis or the next management step in published clinical cases, and on NEJM AI QA, which is drawn from specialist board examinations.
In thinking mode, the gaps to the proprietary models are 9.85 to 11.02 points on JAMA Clinical Challenge and 5.28 to 5.38 points on NEJM AI QA, similar to those for Gemma-4-31B-IT.
The scores of KMed-31B on these clinical case and specialist board questions thus remain close to those of its base model.
With the per-token API price of its base model as a proxy, KMed-31B would extend the price-performance frontier of the comparison set for the six-benchmark average (Figure~\ref{fig:headline-price-pareto}).

\begin{table}[t]
\centering
\small
\caption{Comparison with medical models.}
\label{tab:medical-comparison}
\setlength{\tabcolsep}{3pt}
\renewcommand{\arraystretch}{1.06}
\begin{tabular*}{\linewidth}{@{}l@{\extracolsep{\fill}}ccccccc@{}}
\toprule
Model & Thinking & \begin{tabular}[c]{@{}c@{}}HealthBench\\Hard\end{tabular} & \begin{tabular}[c]{@{}c@{}}HealthBench\\Pro.$^{*}$\end{tabular} & \begin{tabular}[c]{@{}c@{}}MedXpertQA\\Text\end{tabular} & \begin{tabular}[c]{@{}c@{}}JAMA\\C. C.$^{\dagger}$\end{tabular} & \begin{tabular}[c]{@{}c@{}}NEJM\\AI QA\end{tabular} & \begin{tabular}[c]{@{}c@{}}MedQA-\\USMLE\end{tabular} \\
\midrule
\tabitem Cura 1T$^{\ddagger}$~\citep{ai2026cura1tspecializedmodel} & Yes & 36.8\rlap{$^{\mathrm{R}}$}\phantom{0} & \textbf{66.2}\rlap{$^{\mathrm{R}}$}\textbf{\phantom{0}} & \textbf{60.0}\rlap{$^{\mathrm{R}}$}\textbf{\phantom{0}} & n/a & n/a & n/a \\
\tabitem Baichuan-M4~\citep{yang2026baichuanm4clinicalgrademedicalagent} & Yes & 49.7\rlap{$^{\mathrm{R}}$}\phantom{0} & 55.1\rlap{$^{\mathrm{R}}$}\phantom{0} & n/a & n/a & n/a & n/a \\
\tabitem Baichuan-M3 (235B)~\citep{m3team2026baichuanm3modelingclinicalinquiry} & Yes & 44.4\rlap{$^{\mathrm{R}}$}\phantom{0} & 48.56 & 38.74 & 71.03 & 75.15 & 93.57 \\
\tabitem Baichuan-M2 (32B)~\citep{m2team2025baichuanm2scalingmedicalcapability} & Yes & 34.7\rlap{$^{\mathrm{R}}$}\phantom{0} & 46.63 & 27.27 & 65.97 & 68.02 & 89.32 \\
\tabitem Shanzhi-M1 (32B)~\citep{jin2025multidimensionalrubricorientedrewardmodel} & Yes & 44.7\rlap{$^{\mathrm{R}}$}\phantom{0} & 37.88 & 19.32 & 57.81 & 59.68 & 75.43 \\
\tabitem QuarkMed (32B)~\citep{li2025quarkmedmedicalfoundationmodel} & No & n/a & n/a & 28.68\rlap{$^{\mathrm{R}}$} & n/a & n/a & 86.02\rlap{$^{\mathrm{R}}$} \\
\tabitem MedGemma 27B~\citep{sellergren2026medgemmatechnicalreport} & No & 21.79 & 40.56 & 25.7\rlap{$^{\mathrm{R}}$}\phantom{0} & 63.34 & 57.74 & 87.7\rlap{$^{\mathrm{R}}$}\phantom{0} \\
\tabitem MeditronFO 27B~\citep{theimerlienhard2026fullyopenmeditronauditable} & No & 24.07 & 36.20 & 18.00\rlap{$^{\mathrm{R}}$} & 57.01 & 52.12 & 77.61\rlap{$^{\mathrm{R}}$} \\
\tabitem L1-30B-A5B$^{\smash{\P}}$~\citep{lunit2026l130ba5b} & Yes & 24.97 & 39.52 & 31.85 & 64.94 & 70.31 & 87.52 \\
\kmedtablesymbol \textbf{KMed-31B} & Yes & \textbf{51.48} & 59.18 & 58.64 & \textbf{77.86} & \textbf{79.75} & \textbf{94.38} \\
\bottomrule
\end{tabular*}
\vspace{3pt}
\begin{minipage}{\linewidth}
\footnotesize
$^{*}$HealthBench Professional.
$^{\dagger}$JAMA Clinical Challenge.
$^{\ddagger}$Cura 1T uses feedback from HealthBench and MedXpertQA to guide iterative training data refinement and model selection.
$^{\mathrm{R}}$Reported by the model developers, with judges, decoding settings, and numbers of runs that differ across reports and are often unstated.
Other scores are measured with our protocol (Section~\ref{sec:eval-setup}), except that the HealthBench scores of Baichuan-M2, Shanzhi-M1, MedGemma 27B, MeditronFO 27B, and L1-30B-A5B come from a single run.
$^{\P}$Multiple-choice answers are scored by their leading option letter because the model rarely follows the required answer format.
Bold marks the best score in each column.
\end{minipage}
\end{table}

\subsection{Comparison with medical models}
\label{sec:medical-comparison}

Table~\ref{tab:medical-comparison} compares KMed-31B in thinking mode with medical models from other developers.
Scores marked $^{\mathrm{R}}$ are taken from the developers' reports, whose judges, decoding settings, and numbers of runs differ across reports and are often unstated, so comparisons that involve them are less direct than those between scores measured with our protocol.
On the difficult health conversations of HealthBench Hard, KMed-31B scores highest in the table (51.48), above the 49.7 reported for Baichuan-M4.
Among the models from other developers, only the HealthBench Professional (66.2) and MedXpertQA Text (60.0) scores reported for Cura 1T, a closed-weight model, exceed those of KMed-31B.

Six of the compared models have between 27B and 32B parameters, a size comparable to that of KMed-31B: Baichuan-M2, Shanzhi-M1, QuarkMed, MedGemma 27B, MeditronFO 27B, and L1-30B-A5B.
KMed-31B scores higher than all six on every benchmark for which they have a score.
Among the four multiple-choice benchmarks, KMed-31B's margin over the best score among the similarly sized models is largest on MedXpertQA Text at 26.79 points, followed by JAMA Clinical Challenge (11.89), NEJM AI QA (9.44), and MedQA-USMLE (5.06). The corresponding margins are 6.78 points on HealthBench Hard and 12.55 points on HealthBench Professional.
It also scores higher on all six benchmarks than Baichuan-M3, which has 235B parameters.

The medical models from other developers differ in whether they score well on health conversations or on examination questions.
Shanzhi-M1 and Baichuan-M3 report HealthBench Hard scores of 44.7 and 44.4, more than 20 points above Gemma-4-31B-IT in thinking mode (23.80), but score 19.32 and 38.74 on MedXpertQA Text, where Gemma-4-31B-IT scores 58.62.
On the four multiple-choice benchmarks of examination questions and clinical cases, the other developers' models score below Gemma-4-31B-IT wherever scores are available, with the exception of the MedXpertQA Text score reported for Cura 1T.
This pattern is consistent with the cross-task trade-off discussed in Section~\ref{sec:introduction}.
KMed-31B, in contrast, scores highest on HealthBench Hard while its scores on the four multiple-choice benchmarks stay within 0.26 points of Gemma-4-31B-IT.
Comparisons with medical models from other developers do not isolate the effect of our training recipe, as the models were not developed under matched training conditions.
We further examine whether our training recipe generalizes to another model family using Qwen3.8-27B (Section~\ref{sec:architecture-analysis}).
\section{Analysis}
\label{sec:analysis}

\subsection{Comparison of integration methods}
\label{sec:integration-analysis}

Table~\ref{tab:mixed-mopd-summary} compares Mix-RL, MOPD, and MOPD$^2$ as integration methods for obtaining a single non-thinking model with strong performance across all four task categories.
The Multi-Teacher Reference reports the highest score among the four specialists on each benchmark and does not represent a single checkpoint.
The method rows report the percentage of the gap between the original model (Gemma-4-31B-IT) and this reference recovered by each method.
Mix-RL, which trains one policy by RL on all four categories, starts from the clinical inquiry SFT checkpoint and uses 200 training steps, whereas MOPD and MOPD$^2$ start from Gemma-4-31B-IT and use 100 steps.

MOPD$^2$ exceeds the Multi-Teacher Reference on NEJM AI QA and MedQA-USMLE, while recovering 89\% of the reference gain on MedXpertQA Text and 17\% on JAMA Clinical Challenge.
On HealthBench, it recovers 82\% of the reference gain on HealthBench Hard and 57\% on HealthBench Professional.
Compared with MOPD, MOPD$^2$ achieves higher gap recovery on five of the six benchmarks, with similar recovery on HealthBench Hard.
Their six-benchmark average scores are similar, at 65.95 for MOPD and 66.14 for MOPD$^2$.

Mix-RL reaches a comparable average of 65.93, yet achieves a different balance between HealthBench and multiple-choice performance.
Its HealthBench Hard and Professional scores of 43.57 and 58.25 exceed those of MOPD$^2$, but it scores below Gemma-4-31B-IT on three of the four multiple-choice benchmarks.
MOPD$^2$, by comparison, scores above Gemma-4-31B-IT on all six benchmarks.
On MedXpertQA Text and JAMA Clinical Challenge, the SFT initialization of Mix-RL scores 5.61 and 3.16 points below Gemma-4-31B-IT, and Mix-RL recovers 2.33 and 1.77 points of these deficits but remains below Gemma-4-31B-IT.
The similar overall averages thus mask a cross-task trade-off in the final Mix-RL model, consistent with the pattern discussed in Section~\ref{sec:introduction}.

\begin{table}[t]
\centering
\small
\caption{Comparison of integration methods in the non-thinking branch of Gemma-4-31B-IT.
The first two rows report benchmark scores, and the remaining rows report Teacher--Original Performance Gap Recovery (\%).
The Multi-Teacher Reference reports the highest score among the four specialists for each benchmark and does not represent a single checkpoint.}
\label{tab:mixed-mopd-summary}
\setlength{\tabcolsep}{4pt}
\renewcommand{\arraystretch}{1.06}
\begin{tabular*}{\linewidth}{@{}l@{\extracolsep{\fill}}cccccc@{}}
\toprule
Model & \begin{tabular}[c]{@{}c@{}}HealthBench\\Hard\end{tabular} & \begin{tabular}[c]{@{}c@{}}HealthBench\\Pro.$^{*}$\end{tabular} & \begin{tabular}[c]{@{}c@{}}MedXpertQA\\Text\end{tabular} & \begin{tabular}[c]{@{}c@{}}JAMA\\C. C.$^{\dagger}$\end{tabular} & \begin{tabular}[c]{@{}c@{}}NEJM\\AI QA\end{tabular} & \begin{tabular}[c]{@{}c@{}}MedQA-\\USMLE\end{tabular} \\
\midrule
Gemma-4-31B-IT & 22.95 & 37.92 & 53.86 & 75.89 & 75.54 & 93.08 \\
Multi-Teacher Reference & 46.04 & 59.62 & 56.63 & 76.91 & 77.17 & 94.34 \\
\midrule
\multicolumn{7}{@{}l@{}}{\textit{Teacher--Original Performance Gap Recovery (\%)}} \\
Mix-RL & 89\% & 94\% & \textcolor{red}{$-$118\%} & \textcolor{red}{$-$136\%} & 9\% & \textcolor{red}{$-$6\%} \\
MOPD & 82\% & 55\% & 81\% & 7\% & \textcolor{ForestGreen}{125\%} & 85\% \\
MOPD$^2$ & 82\% & 57\% & 89\% & 17\% & \textcolor{ForestGreen}{153\%} & \textcolor{ForestGreen}{103\%} \\
\bottomrule
\end{tabular*}
\vspace{3pt}
\begin{minipage}{\linewidth}
\footnotesize
$^{*}$HealthBench Professional.
$^{\dagger}$JAMA Clinical Challenge.
Mix-RL initializes from the clinical inquiry SFT checkpoint, whereas MOPD and MOPD$^2$ initialize from Gemma-4-31B-IT.
\end{minipage}
\end{table}

\subsection{Generalization to another model family}
\label{sec:architecture-analysis}

To assess whether our medical post-training recipe generalizes beyond the Gemma family, we apply the same specialist training and integration procedure to Qwen3.8-27B in thinking mode (Table~\ref{tab:qwen-Mix-RL}).
Since both the base model and the response mode differ from the non-thinking Gemma-4-31B-IT run, this experiment does not isolate the effect of model family.

The resulting MOPD$^2$ model scores above Qwen3.8-27B on all six medical benchmarks.
MOPD$^2$ scores above the Multi-Teacher Reference on all four multiple-choice benchmarks, by $+0.10$ to $+1.62$ points, and below it on HealthBench Hard ($-1.25$) and HealthBench Professional ($-6.01$).
Measured from Qwen3.8-27B, it recovers 92\% of the reference gain on HealthBench Hard and 61\% on HealthBench Professional, compared with 82\% and 57\% for Gemma-4-31B-IT in non-thinking mode (Section~\ref{sec:integration-analysis}).
In both settings, gap recovery is lower on HealthBench Professional than on HealthBench Hard.

Compared with MOPD, MOPD$^2$ scores higher on both HealthBench benchmarks (50.14 against 47.28 for Hard and 56.28 against 55.28 for Professional).
It also achieves higher gap recovery than MOPD on three of the four multiple-choice benchmarks, including an increase from 96\% to 116\% on MedXpertQA Text.
The exception is JAMA Clinical Challenge, where recovery decreases from 121\% to 113\%.
Mix-RL scores above Qwen3.8-27B on both HealthBench benchmarks (47.77 on Hard and 60.22 on Professional) and falls below Qwen3.8-27B only on MedXpertQA Text, by 0.90 points.
Unlike in the non-thinking Gemma-4-31B-IT run, MOPD$^2$ scores higher than Mix-RL on HealthBench Hard, while Mix-RL remains higher on HealthBench Professional.

\begin{table}[t]
\centering
\small
\caption{Comparison of integration methods for Qwen3.8-27B in thinking mode.
The first two rows report benchmark scores, and the remaining rows report Teacher--Original Performance Gap Recovery (\%).
The Multi-Teacher Reference reports the highest score among the four specialists for each benchmark and does not represent a single checkpoint.}
\label{tab:qwen-Mix-RL}
\renewcommand{\arraystretch}{1.06}
\setlength{\tabcolsep}{4pt}
\begin{tabular*}{\linewidth}{@{}l@{\extracolsep{\fill}}cccccc@{}}
\toprule
Model & \begin{tabular}[c]{@{}c@{}}HealthBench\\Hard\end{tabular} & \begin{tabular}[c]{@{}c@{}}HealthBench\\Pro.$^{*}$\end{tabular} & \begin{tabular}[c]{@{}c@{}}MedXpertQA\\Text\end{tabular} & \begin{tabular}[c]{@{}c@{}}JAMA\\C. C.$^{\dagger}$\end{tabular} & \begin{tabular}[c]{@{}c@{}}NEJM\\AI QA\end{tabular} & \begin{tabular}[c]{@{}c@{}}MedQA-\\USMLE\end{tabular} \\
\midrule
Qwen3.8-27B & 35.64 & 46.90 & 43.95 & 77.94 & 75.45 & 94.27 \\
Multi-Teacher Reference & 51.39 & 62.29 & 45.95 & 80.72 & 77.44 & 95.26 \\
\midrule
\multicolumn{7}{@{}l@{}}{\textit{Teacher--Original Performance Gap Recovery (\%)}} \\
Mix-RL & 77\% & 87\% & \textcolor{red}{$-$45\%} & 89\% & 49\% & 16\% \\
MOPD & 74\% & 54\% & 96\% & \textcolor{ForestGreen}{121\%} & \textcolor{ForestGreen}{154\%} & 86\% \\
MOPD$^2$ & 92\% & 61\% & \textcolor{ForestGreen}{116\%} & \textcolor{ForestGreen}{113\%} & \textcolor{ForestGreen}{181\%} & \textcolor{ForestGreen}{110\%} \\
\bottomrule
\end{tabular*}
\vspace{3pt}
\begin{minipage}{\linewidth}
\footnotesize
$^{*}$HealthBench Professional.
$^{\dagger}$JAMA Clinical Challenge.
Mix-RL initializes from the clinical inquiry SFT checkpoint.
\end{minipage}
\end{table}

\subsection{Retention of general capabilities}
\label{sec:general-retention}

\begin{table}[t]
\centering
\small
\caption{General benchmark performance before and after medical post-training.}
\label{tab:ood-retention}
\renewcommand{\arraystretch}{1.06}
\setlength{\tabcolsep}{4pt}
\begin{tabular*}{\linewidth}{@{}l@{\extracolsep{\fill}}cccccc@{}}
\toprule
Model & MATH-500 & \begin{tabular}[c]{@{}c@{}}GPQA\\Diamond\end{tabular} & $\tau$-bench & BFCL-v3 & IFEval & MMLU-Pro \\
\midrule
\multicolumn{7}{@{}l@{}}{\textit{Gemma-4, non-thinking mode}} \\
\tabitem Gemma-4-31B-IT & 97.40 & 80.30 & 70.50 & 74.96 & 90.94 & 85.12 \\
\kmedtablesymbol KMed-31B & 98.00 & 83.50 & 69.50 & 72.33 & 88.23 & 85.94 \\
\midrule
\multicolumn{7}{@{}l@{}}{\textit{Qwen3.8, thinking mode}} \\
\tabitem Qwen3.8-27B & 97.52 & 81.26 & 70.97 & 70.84 & 91.77 & 85.18 \\
\tabitem Qwen3.8-27B (MOPD$^2$) & 98.40 & 89.73 & 71.52 & 71.82 & 91.93 & 86.10 \\
\bottomrule
\end{tabular*}
\vspace{3pt}
\begin{minipage}{\linewidth}
\footnotesize
All results are avg@10.
IFEval reports prompt-level strict accuracy; $\tau$-bench reports pass$^{1}$; the other benchmarks report accuracy.
\end{minipage}
\end{table}

We also evaluate whether the gains on medical benchmarks are accompanied by changes in general capabilities, using six general benchmarks (Table~\ref{tab:ood-retention}).
These include MATH-500~\citep{hendrycks2021measuringmathematicalproblemsolving,lightman2023letsverifystepstep} for mathematics, GPQA Diamond~\citep{rein2023gpqagraduatelevelgoogleproof} for graduate-level science questions, and MMLU-Pro~\citep{wang2024mmluprorobustchallengingmultitask} for broad knowledge.
We use $\tau$-bench~\citep{yao2024taubenchbenchmarktoolagentuser} to evaluate tool use in conversations with simulated users, BFCL-v3~\citep{patil2025bfcl} for function calling, and IFEval~\citep{zhou2023instructionfollowingevaluationlargelanguage} for instruction following.

Relative to Gemma-4-31B-IT, the non-thinking KMed-31B model scores 2.71 points lower on IFEval, 2.63 lower on BFCL-v3, and 1.00 lower on $\tau$-bench, but 3.20 points higher on GPQA Diamond, 0.82 higher on MMLU-Pro, and 0.60 higher on MATH-500.
For Qwen3.8-27B in thinking mode, the MOPD$^2$ model scores higher than the base model on all six benchmarks, with the largest increase on GPQA Diamond (81.26 to 89.73).
Performance on the evaluated general benchmarks is thus largely preserved in both settings.

\subsection{Gains in thinking and non-thinking modes}
\label{sec:thinking-gap}

The separately trained thinking and non-thinking KMed-31B models differ in their gains over Gemma-4-31B-IT: the thinking model improves more on HealthBench, whereas the non-thinking model improves more on all four multiple-choice benchmarks (Table~\ref{tab:general-comparison}).
On HealthBench Hard, the thinking model gains 27.68 points over its base model, compared with 18.92 points for the non-thinking model.
The corresponding gains on HealthBench Professional are 17.21 and 12.28 points.
The gap between thinking and non-thinking scores therefore widens from 0.85 points in Gemma-4-31B-IT to 9.61 points in KMed-31B on HealthBench Hard, and from 4.05 to 8.98 points on HealthBench Professional.
On the four multiple-choice benchmarks, however, the changes in the thinking model remain small, ranging from $-0.16$ to $+0.26$ points, while the non-thinking model improves on all four benchmarks by $+0.17$ to $+2.49$ points.

\subsection{Gains by HealthBench theme and rubric axis}
\label{sec:hb-breakdown}

\begin{table}[t]
\centering
\small
\caption{HealthBench breakdown of Gemma-4-31B-IT and KMed-31B (non-thinking, avg@3).
Theme, use-case, and query-type scores average the example scores within each tag, and axis scores average the per-example rubric-axis scores.}
\label{tab:hb-breakdown}
\setlength{\tabcolsep}{6pt}
\begin{tabular}{@{}lrrr@{}}
\toprule
Group & Gemma-4-31B-IT & \kmedtablesymbol KMed-31B & $\Delta$ \\
\midrule
\multicolumn{4}{@{}l@{}}{\textit{HealthBench Hard, theme}} \\
\tabitem Hedging & 25.09 & 50.07 & +24.98 \\
\tabitem Context seeking & 22.90 & 47.32 & +24.42 \\
\tabitem Emergency referrals & 31.64 & 53.30 & +21.66 \\
\tabitem Global health & 23.93 & 42.99 & +19.06 \\
\tabitem Health data tasks & 19.10 & 32.21 & +13.11 \\
\tabitem Communication & 19.16 & 31.92 & +12.76 \\
\tabitem Complex responses & 18.81 & 27.29 & +8.48 \\
\addlinespace[3pt]
\multicolumn{4}{@{}l@{}}{\textit{HealthBench Hard, rubric axis}} \\
\tabitem Context awareness & 8.95 & 36.22 & +27.27 \\
\tabitem Completeness & 21.41 & 45.79 & +24.38 \\
\tabitem Accuracy & 34.90 & 43.98 & +9.08 \\
\tabitem Instruction following & 55.37 & 58.80 & +3.43 \\
\tabitem Communication quality & 72.95 & 69.79 & $-$3.16 \\
\midrule
\multicolumn{4}{@{}l@{}}{\textit{HealthBench Professional, use case}} \\
\tabitem Writing & 20.94 & 41.70 & +20.76 \\
\tabitem Consult & 37.81 & 49.40 & +11.59 \\
\tabitem Research & 54.51 & 59.69 & +5.18 \\
\addlinespace[3pt]
\multicolumn{4}{@{}l@{}}{\textit{HealthBench Professional, query type}} \\
\tabitem Red teaming & 9.61 & 30.75 & +21.14 \\
\tabitem Good faith & 54.11 & 61.32 & +7.21 \\
\midrule
Overall (Hard) & 22.95 & 41.87 & +18.92 \\
Overall (Professional) & 37.92 & 50.20 & +12.28 \\
\bottomrule
\end{tabular}
\end{table}

To examine which aspects of HealthBench performance improve most, we compare Gemma-4-31B-IT and KMed-31B in non-thinking mode by theme and rubric axis on HealthBench Hard, and by use case and query type on HealthBench Professional (Table~\ref{tab:hb-breakdown}).
On HealthBench Hard, the largest increases are on the context awareness and completeness axes (8.95 to 36.22 and 21.41 to 45.79), and on the hedging and context-seeking themes (25.09 to 50.07 and 22.90 to 47.32), whereas communication quality decreases slightly (72.95 to 69.79).
The accuracy axis also improves (34.90 to 43.98), although less than context awareness and completeness.
On HealthBench Professional, the largest increases are for writing among use cases (20.94 to 41.70) and red-teaming among query types (9.61 to 30.75).

The gains in context awareness, hedging, and context seeking are consistent with the behaviors targeted by the clinical inquiry data.
SHD includes conversations in which some details of the user's situation are initially withheld from the assistant, and the SFT responses are generated with the context-and-premise prompt, which asks the generator to identify missing information, ask focused questions, state assumptions, and check the premises of the request (Section~\ref{sec:synthesis}).

\subsection{Training stages of the clinical inquiry specialist}
\label{sec:ci-stages}

\begin{figure}[t]
\centering
\includegraphics[width=0.85\linewidth]{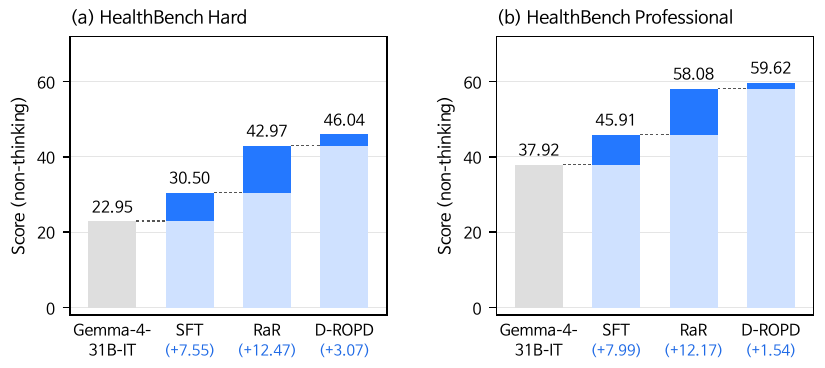}
\caption{HealthBench performance across the training stages of the clinical inquiry specialist, evaluated in non-thinking mode.
Scores rise after SFT, RL with fixed rubrics (RaR), and RL with online rubrics (D-ROPD).
Scores are reported without length adjustment, and the annotations show the changes from the preceding stage.}
\label{fig:clinical-inquiry-stages}
\end{figure}

We compare the gains across SFT, RaR, and D-ROPD using the unadjusted HealthBench scores of the non-thinking clinical inquiry specialist before multi-teacher integration (Figure~\ref{fig:clinical-inquiry-stages}).
The HealthBench Hard score increases from 22.95 to 46.04, and the HealthBench Professional score increases from 37.92 to 59.62.
SFT yields similar gains on the two benchmarks, improving HealthBench Hard by 7.55 points and HealthBench Professional by 7.99 points.
RaR, which uses fixed rubrics, produces the largest stage-to-stage increases: 12.47 and 12.17 points, respectively.
D-ROPD, which uses online rubrics, further improves the scores by 3.07 points on HealthBench Hard and 1.54 points on HealthBench Professional.

\subsection{Explicit and implicit reward aggregation}
\label{sec:rar-aggregation}

\begin{figure}[t]
\centering
\includegraphics[width=0.85\linewidth]{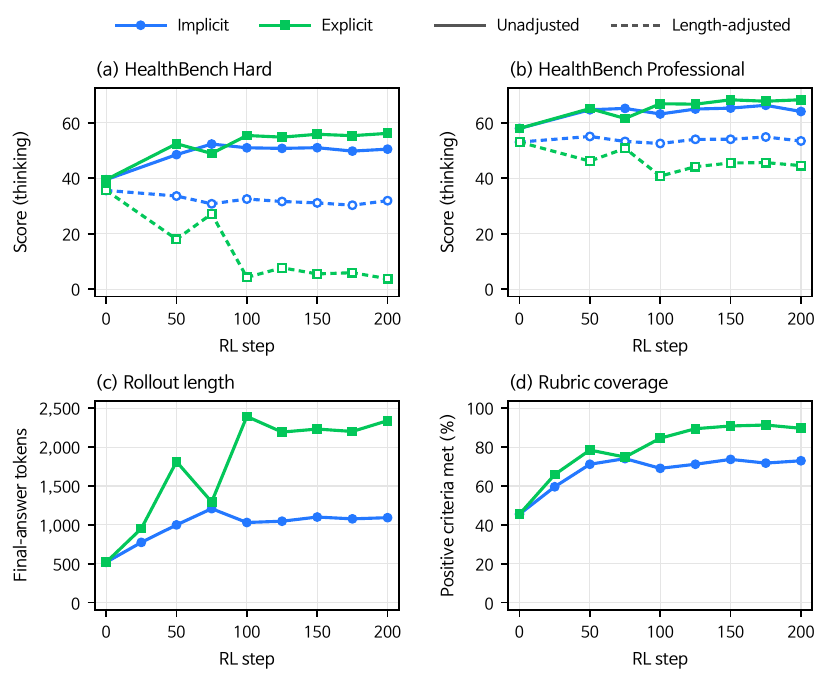}
\caption{Implicit and explicit reward aggregation during clinical inquiry RL.
The top panels show unadjusted and length-adjusted scores on HealthBench Hard and HealthBench Professional; the bottom panels show final-answer length and the proportion of positive rubric criteria met.}
\label{fig:rar-aggregation}
\end{figure}

Figure~\ref{fig:rar-aggregation} compares two ways of computing rubric-based rewards during the RaR stage of the thinking-mode clinical inquiry specialist~\citep{gunjal2025rubricsrewardsreinforcementlearning}.
Explicit aggregation scores each rubric criterion separately and combines the verdicts into a weighted reward, while implicit aggregation asks the judge to assign a single score using the whole rubric.
After 200 steps, explicit aggregation gives higher unadjusted scores on HealthBench Hard and HealthBench Professional, 56.19 and 68.30 against 50.50 and 64.07 with implicit aggregation.
However, explicit aggregation yields lower length-adjusted scores: 3.82 and 44.57, compared with 31.94 and 53.45 for implicit aggregation.
During training, the final answers under explicit aggregation grow to about twice the length of those under implicit aggregation, and the share of positive rubric criteria met by the rollouts reaches 89.6\% against 72.9\% at step 200.
One possible explanation is that implicit aggregation stops raising the reward once further criteria no longer change the judge's overall score, whereas explicit aggregation keeps rewarding each remaining criterion, so responses grow as more criteria are covered.
A recent study reports a similar pattern in medical and dialogue tasks, where scoring the same criteria with implicit aggregation preserved response appropriateness better than additive aggregation~\citep{liu2026scoringhigheransweringworse}.
We therefore use implicit aggregation in the RaR stage.

\section{Limitations}
\label{sec:limitations}

Although the recipe improves HealthBench scores in both response modes and on two base models without more than a marginal decrease in the multiple-choice medical scores, our work has limitations, several of which we note here.
Our task categories and medical benchmarks cover examination-style questions, diagnosis from case reports, and consultation dialogue, but not tasks on electronic health records, such as summarizing clinical notes or answering questions over structured patient data, or agentic tasks in which a model takes several steps with tools in a clinical environment~\citep{jiang2025medagentbenchrealisticvirtualehr}.
The recipe trains a separate specialist for each category, so such tasks could be added as new categories with suitable rewards, and we leave this extension to future work.
In addition, our HealthBench scores are graded by GPT-5-mini rather than by the graders of the original reports, and unadjusted rubric scores tend to favor longer responses.
Under length adjustment, the ordering on HealthBench Hard changes (Appendix~\ref{app:length-adjusted}), and keeping the gains with shorter responses is also left to future work.
Some comparisons in Section~\ref{sec:analysis} differ in initialization and training length, and most configurations were trained once, so small differences between them should be interpreted with caution.
More broadly, our evaluation relies on benchmarks rather than studies in clinical settings, so the results do not show how KMed-31B would perform in patient care, where its outputs would require review by clinicians.
\section{Related Work}
\label{sec:related}

\paragraph{Medical language models.}
Most recent medical models come with technical reports that describe their post-training recipes.
Baichuan-M1 is pretrained for medicine from scratch~\citep{wang2025baichuanm1pushingmedicalcapability}, while most later reports adapt a general base model.
Baichuan-M2 runs RL against a verifier system made up of a patient simulator and a clinical rubrics generator~\citep{m2team2025baichuanm2scalingmedicalcapability}, Baichuan-M3 models clinical inquiry and targets hallucination~\citep{m3team2026baichuanm3modelingclinicalinquiry}, and Baichuan-M4 extends the model into an agent system for continuous care with patient memory~\citep{yang2026baichuanm4clinicalgrademedicalagent}.
QuarkMed combines data curation, medical retrieval augmentation, and large-scale verifiable RL~\citep{li2025quarkmedmedicalfoundationmodel}.
MedGemma and MedGemma 1.5 adapt Gemma 3 to medical text and imaging~\citep{sellergren2026medgemmatechnicalreport,sellergren2026medgemma15technicalreport}, and Cura 1T post-trains a one-trillion-parameter model through a human-gated self-improvement loop~\citep{ai2026cura1tspecializedmodel}.
Other medical models include Shanzhi-M1~\citep{jin2025multidimensionalrubricorientedrewardmodel}, MeditronFO 27B~\citep{theimerlienhard2026fullyopenmeditronauditable}, and L1-30B-A5B~\citep{lunit2026l130ba5b}.
The report closest to ours is Baichuan-M3, which trains medical experts with independent RL from a shared initialization and integrates them through multi-stage distillation that ends in MOPD.
Its experts are organized by clinical capability (clinical inquiry and healthcare consultation, plus a generalist), whereas our specialists follow task categories that differ in the expected output and in the available training feedback (Section~\ref{sec:taxonomy}).

\paragraph{Clinical interaction and simulation.}
Flan-PaLM was the first model to exceed the commonly quoted pass mark on questions in the style of the U.S. medical licensing examination, with 67.6\% on MedQA-USMLE, and Med-PaLM added physician evaluation of long-form answers~\citep{singhal2022largelanguagemodelsencode,singhal2023expertlevelmedicalquestionanswering}.
Med-PaLM 2 then reached up to 86.5\% on the same questions~\citep{singhal2023expertlevelmedicalquestionanswering}.
AMIE optimizes a model for diagnostic dialogue through self-play between a simulated patient, a doctor agent, and a critic.
In a randomized remote objective structured clinical examination (OSCE) against 20 primary care physicians, specialists rated it higher on 30 of 32 axes~\citep{tu2024conversationaldiagnosticai}.
MAI-DxO prompts an off-the-shelf model to act as a panel of physicians on 304 NEJM clinicopathological cases converted into sequential encounters, and achieves higher diagnostic accuracy than practicing generalist physicians~\citep{nori2025sequentialdiagnosislanguagemodels}.
Dr.~CaBot generates expert-style diagnostic presentations from NEJM case descriptions~\citep{buckley2026teachinglargelanguagemodels}.
AgentClinic and MedAgentBench evaluate models in simulated clinics and in a virtual electronic health record environment~\citep{schmidgall2025agentclinicmultimodalagentbenchmark,jiang2025medagentbenchrealisticvirtualehr}.
Our SHD simulator separates what the user knows from what the assistant can see, as patient simulators for diagnostic dialogue do, but uses the resulting conversations as training queries for single assistant responses rather than to evaluate a model over an interactive encounter (Section~\ref{sec:synthesis}).

\paragraph{Rewards for open-ended medical responses.}
When an open-ended response cannot be scored against a single reference answer, as is common in clinical inquiry, RL often relies on LLM judges for its reward.
Rubrics as Rewards scores responses against instance-specific rubrics and reports a relative improvement of up to 31\% on HealthBench over judges that assign a direct Likert score~\citep{gunjal2025rubricsrewardsreinforcementlearning}, and Rubric Anchors scales rubric rewards to more than 10,000 rubrics for open-ended tasks~\citep{huang2025reinforcementlearningrubricanchors}.
Fixed rubrics can be updated during training: OnlineRubrics adds criteria elicited from pairwise comparisons of current and reference policy responses~\citep{rezaei2025onlinerubrics}, and ROPD induces prompt-specific rubrics from contrasts between teacher and student responses and uses them to score the student's rollouts~\citep{fang2026rubricbasedonpolicydistillation}.
ConRub-Med applies RL with consensus rubrics to open-ended medical question answering and reports that rubric-based RL raises HealthBench Hard while lowering multiple-choice accuracy~\citep{zhu2026conrubmedreinforcementlearningconsensus}.
Rubric rewards can also be exploited: in medical and science domains, weak verifiers produce proxy-reward gains that do not transfer to reference judges, and even strong rubric-based verifiers can prefer responses that rubric-free judges rate lower~\citep{mahmoud2026rewardhackingrubricbasedreinforcement}.
Gold reference answers are another reward source for non-verifiable tasks~\citep{shi2026referencesimprovellmalignment,jiang2026verifiabledotrewardchain}.
Our clinical inquiry specialist uses rubric rewards in two stages, first with a rubric written in advance for each query and then with a rubric generated during training for each sampled query from fixed reference responses of four different models and the query's current rollouts, which we call D-ROPD (Section~\ref{sec:specialist-teachers}).

\paragraph{Interference in multi-domain post-training.}
Different domains are trained with different reward pipelines, and combining them in one model is a recurring problem in post-training.
Mix-RL pools domain data into a single run and depends on the mixture proportions~\citep{wu2026consolidatingrlvrcapabilitiesdomains}, while cascaded RL trains domains in sequence~\citep{wang2026nemotroncascadescalingcascadedreinforcement}.
Both strategies have been reported to show interference between domains.
In Mix-RL, some tasks produce much larger gradients than others without greater learning gains~\citep{wu2025imbalancedgradientsrlposttraining}, and cross-domain interference appears at the behavioral and gradient levels under both strategies~\citep{cai2026advancinggeneralpurposereasoningmodels}.
Proposed remedies operate within a single policy, through module-level gradient surgery~\citep{cai2026advancinggeneralpurposereasoningmodels}, projection of later updates onto capability directions from earlier stages~\citep{zhang2026carerlcapabilityawarereinforcementlearning}, task scheduling by backward transfer~\citep{li2025omnithinkerscalingmultitaskrl}, or a short refresh of an earlier domain~\citep{yang2026localperturbationtheorycrossdomain}.
We treat medical task categories that differ in expected output and available feedback as such domains and integrate their specialists rather than training a single policy jointly on all categories (Sections~\ref{sec:taxonomy} and~\ref{sec:mopd2}).

\paragraph{Multi-teacher on-policy distillation.}
MOPD instead trains one RL teacher per domain and distills all teachers into a student on the student's own rollouts; on Qwen3-30B-A3B it outperforms Mix-RL, cascaded RL, off-policy fine-tuning, and parameter merging~\citep{ma2026mopdmultiteacheronpolicydistillation}.
Nemotron-Cascade 2 applies multi-domain on-policy distillation from intermediate teachers to recover regressions during cascaded RL~\citep{yang2026nemotroncascade2posttrainingllms}.
Follow-up work changes the distillation target or the schedule.
OPD$^2$ distills the difference between a teacher and its base model prior to instruction tuning for reasoning~\citep{heo2026onpolicydeltadistillation}, D$^3$-MOPD adapts the domain mixture online from per-domain reverse-KL trajectories~\citep{sun2026d3mopddynamicdomainscheduling}, Open-MOPD traces the gap to an oracle ensemble to how the token-level optimization budget is allocated across domains~\citep{gao2026openmopddiagnosingfixingcapability}, and SMOPD trains one teacher per reward and merges them by distillation~\citep{wang2026smopdmultirewardreinforcementlearning}.
Experts can also be combined in parameter space with task vectors~\citep{ilharco2023editingmodelstaskarithmetic} or with per-sample weight interpolation~\citep{oh2025dawintrainingfreedynamicweight}.
Comparing merging, Mix-RL, and MOPD with shared experts, \citet{wu2026consolidatingrlvrcapabilitiesdomains} find that average scores differ by at most 1.4 points, while single benchmarks differ by up to 8.6 points.
MOPD$^2$ keeps the per-category specialists of MOPD as teachers and replaces their token signal with the delta of OPD$^2$ between each specialist and a pretrained reference model (Section~\ref{sec:mopd2}).
\section{Conclusion}
\label{sec:conclusion}

We presented KMed-31B, a medical model post-trained from Gemma-4-31B-IT.
Its recipe trains one specialist for each medical task category with a reward that the category's feedback supports, builds a synthetic corpus of consultation dialogues for clinical inquiry, and integrates the specialists into a single student with MOPD$^2$.
In both non-thinking and thinking modes, KMed-31B has the highest unadjusted HealthBench Hard score among the compared models, including proprietary ones, and scores higher on HealthBench Professional than every open-weight model we compare, while none of its multiple-choice medical scores falls more than marginally below that of the base model.
Applied to Qwen3.8-27B, the same recipe raises both HealthBench scores, which suggests that it is not specific to one base model.
These results indicate that training specialists for medical task categories defined by expected output and available training feedback, followed by distillation into a single model, is a practical way to improve responses in clinical conversations while preserving performance on examination-style questions, at a model size that is more feasible for on-premises deployment.
We plan to release an open-weight version of KMed in the near future.
\section*{Contributions}

Within each role, names are listed in alphabetical order by first name, followed by the last name.

\paragraph{Model Planning, Research, Training, and Evaluation.}
Byeongho Heo, Dongyoon Han, Jaedeok Lee, Sangdoo Yun, Seongsu Bae, Yera Choi

\paragraph{Technical Writing.}
Byeongho Heo, Dongyoon Han, Jaedeok Lee, Keonwoo Kim, Sangdoo Yun, Seongsu Bae, Yera Choi

\paragraph{Data and Model Serving.}
Jaedeok Lee, Seongsu Bae, Yera Choi

\paragraph{Downstream Evaluation and Feedback.}
Heeyoung Kwak, Hyejeong Jo, Hyeonsoo Lee, Hyeseon Ko, Keonwoo Kim, Kilyong Shin, Sewon Kim

\paragraph{Software Engineering, Planning, and Quality Assurance.}
Geonho Kim, Hana Kang, Hyun Ah Kim, Jina You, Jinhong Kim, Sangwoong Lee, Seolim Oh, Soojin Kim, Zoo Hyun Lee

\paragraph{Clinical Advisory.}
Dongchul Cha, Hyeonhoon Lee$^\dagger$, Hyung-Chul Lee$^\dagger$, Yunwook Sung$^\ddagger$

\paragraph{Project Director.}
Haanju Yoo

\medskip
\noindent
$^\dagger$Healthcare AI Research Institute, Seoul National University Hospital.\\
$^\ddagger$Work done while at the Healthcare AI Research Institute, Seoul National University Hospital.

{\small
\bibliographystyle{unsrtnat}
\bibliography{kmed}
}
\appendix

\clearpage
\section{Appendix}
\subsection{Training hyperparameters}
\label{app:hyperparameters}

\begin{table}[htbp]
\centering
\small
\caption{Training hyperparameters of the non-thinking Gemma-4-31B-IT pipeline.
The RLVR specialists are the MedQA-OE, MedMCQA, and MedCaseReasoning teachers; the clinical inquiry teacher is trained with SFT, RaR, and D-ROPD.}
\label{tab:hyperparameters}
\setlength{\tabcolsep}{4pt}
\begin{tabular*}{\linewidth}{@{\extracolsep{\fill}}lccccc@{}}
\toprule
 & & \multicolumn{3}{c}{Clinical inquiry teacher} & \\
\cmidrule(lr){3-5}
Configuration & \begin{tabular}[c]{@{}c@{}}RLVR\\specialists\end{tabular} & SFT & RaR & D-ROPD & MOPD$^2$ \\
\midrule
Algorithm        & GRPO$^{a}$ & SFT & GRPO$^{b}$ & D-ROPD$^{c}$ & MOPD$^2$ \\
Initial checkpoint       & Gemma-4-31B-IT & Gemma-4-31B-IT & SFT & RaR (step 200) & Gemma-4-31B-IT \\
\midrule
Learning rate    & $2\times10^{-6}$ & $1\times10^{-5}$ & $2\times10^{-6}$ & $3\times10^{-6}$ & $5\times10^{-6}$ \\
Warm-up steps      & 20 & 0 & 20 & 5 & 10 \\
LR schedule      & Cosine & Cosine & Cosine & Cosine & Constant \\
Minimum LR       & $2\times10^{-7}$ & -- & $2\times10^{-7}$ & $3\times10^{-7}$ & -- \\
\midrule
Batch size       & $36\times16$ & 120 & $36\times16$ & $36\times16$ & $512\times1$ \\
KL coefficient   & $0$ / $10^{-2}$\,$^{d}$ & -- & $10^{-2}$ & $10^{-2}$ & $5\times10^{-4}$ \\
Sequence limit   & 14k (12k) & 8k & 14k (12k) & 14k (12k) & 11k (8k) \\
Training length  & 200 steps & 5 epochs & 200 steps & 50 steps & 100 steps \\
Truncated IS ratio & 2 & -- & 2 & 2 & 2 \\
\midrule
Other            & {\footnotesize Dynamic sampling} & -- & -- & -- & {\footnotesize Rollout $T$ 1.2} \\
\bottomrule
\end{tabular*}
\vspace{3pt}
\begin{minipage}{\linewidth}
\footnotesize
$^{a}$Verifiable answer reward.
$^{b}$Implicit rubric reward.
$^{c}$Rubrics generated from multiple teachers' reference responses and the current policy responses.
$^{d}$0 for MedQA-OE and MedMCQA; $10^{-2}$ for MedCaseReasoning.
LR schedule: decay shape after the warm-up steps.
Batch size: sequences for SFT; prompts $\times$ rollouts per prompt for the other stages.
Sequence limit: total tokens, with the generation budget in parentheses (k = 1,024 tokens).
IS: importance sampling.
$T$: rollout temperature.
\end{minipage}
\end{table}

\subsection{Length-adjusted HealthBench scores}
\label{app:length-adjusted}

\begin{table}[htbp]
\centering
\small
\caption{Unadjusted and length-adjusted HealthBench scores in non-thinking mode.}
\label{tab:length-adjusted}
\setlength{\tabcolsep}{5pt}
\begin{tabular*}{\linewidth}{@{}l@{\extracolsep{\fill}}rrrr@{}}
\toprule
& \multicolumn{2}{c}{HealthBench Hard} & \multicolumn{2}{c}{HealthBench Professional} \\
\cmidrule(lr){2-3}\cmidrule(l){4-5}
Model & Unadjusted & Length-adjusted & Unadjusted & Length-adjusted \\
\midrule
Claude-Opus-5 & 38.29 & 31.97 & 67.94 & 57.76 \\
GPT-5.6-sol & 38.59 & 41.21 & 62.36 & 59.08 \\
Gemma-4-31B-IT & 22.95 & 18.50 & 37.92 & 34.17 \\
\kmedtablesymbol KMed-31B & 41.87 & 25.25 & 50.20 & 43.10 \\
\bottomrule
\end{tabular*}
\vspace{3pt}
\begin{minipage}{\linewidth}
\footnotesize
All scores are avg@3.
\end{minipage}
\end{table}

Since HealthBench scores can be sensitive to response length, we also report length-adjusted scores using the procedure introduced for HealthBench Professional~\citep{hicks2026healthbenchprofessional}.
For each example, let $s$ denote the unadjusted rubric score before multiplication by 100 and $\ell$ the final-answer length in characters, excluding thinking tokens.
The adjusted score is $s_{\mathrm{len}} = s - c(\ell - 2000)$, so a 2,000-character answer receives no adjustment.
We use $c = 2.94 \times 10^{-5}$ for HealthBench Professional and $c = 7.84 \times 10^{-5}$ for HealthBench Hard~\citep{openai2026gpt6octoberupdate}.
We average the adjusted example scores, clip the mean to $[0,1]$, and multiply by 100 for reporting.

Table~\ref{tab:length-adjusted} reports the unadjusted and length-adjusted scores in non-thinking mode.
After adjustment, KMed-31B scores 25.25 on HealthBench Hard and 43.10 on HealthBench Professional, compared with 41.21 and 59.08 for GPT-5.6-sol and 31.97 and 57.76 for Claude-Opus-5, respectively.
This changes the relative ranking of KMed-31B and both proprietary models on HealthBench Hard compared with the unadjusted results in Section~\ref{sec:general-comparison}.
The differences between the unadjusted and adjusted scores imply mean final-answer lengths of approximately 4,100 and 4,400 characters for KMed-31B on HealthBench Hard and HealthBench Professional, respectively, compared with 1,700 and 3,100 characters for GPT-5.6-sol.
Our post-training does not explicitly constrain response length, and improving concision while preserving the gains in unadjusted scores remains a direction for future work.

\subsection{Context-and-premise prompt}
\label{app:context-premise-prompt}

Clinical inquiry requests may omit information that materially affects the answer or rest on a mistaken premise.
The context-and-premise prompt below encourages the generator to address these gaps and errors while providing useful guidance under explicit assumptions.
In an auxiliary evaluation on HealthBench Professional, all four response generators obtained higher scores with the prompt than without it, both before and after length adjustment.
This observation supports its use in generating responses for clinical inquiry training: the SFT demonstrations provide direct supervision (Section~\ref{sec:synthesis}), while the D-ROPD reference responses inform rubric construction and verification (Section~\ref{sec:specialist-teachers}).

\begin{tcolorbox}[
  colback=white,
  colframe=black!75,
  colbacktitle=black!75,
  coltitle=white,
  fonttitle=\normalfont\bfseries,
  title={Context-and-premise prompt},
  rounded corners,
  arc=1.5mm,
  boxrule=0.5pt,
  width=\linewidth
]
\small\ttfamily\raggedright
\setlength{\parindent}{0pt}
\setlength{\parskip}{0.7\baselineskip}
You are a helpful assistant.

When missing information could materially change the answer, identify it and ask focused questions when needed.
Provide useful guidance with explicit assumptions where possible.
Do not invent patient-specific facts.

Check the premises of the request for consequential errors or inconsistencies, including when drafting, translating, or summarizing.
Flag important issues separately from the requested output, and retain essential qualifications even when asked to be brief.
\end{tcolorbox}

\end{document}